\documentclass[twoside]{article}
\usepackage{ecj,palatino,epsfig,latexsym,natbib}

\usepackage{graphicx}
\usepackage{subcaption}
\usepackage{hyperref}
\usepackage{textcomp}
\usepackage{verbatim}
\usepackage{tabularx}
\usepackage{algorithm}
\usepackage{algpseudocode}
\usepackage{amsmath}
\newcommand{\pluseq}{\mathrel{+}=}

\begin{document}

	\ecjHeader{x}{x}{xxx-xxx}{201X}{Guiding Neuroevolution with Structural Objectives}{K.O Ellefsen, J. Huizinga and J. Torresen}
	\title{\bf Guiding Neuroevolution with Structural Objectives}  
	
	\author{\name{\bf Kai Olav Ellefsen} \hfill \addr{kaiolae@ifi.uio.no}\\ 
		\addr{Department of Informatics, University of Oslo, Norway}		
		\AND
		\name{\bf Joost Huizinga} \hfill \addr{jhuizinga@uber.com}\\
		\addr{Uber AI Labs, USA}
		\AND
		\name{\bf Jim Torresen} \hfill \addr{jimtoer@ifi.uio.no}\\
		\addr{Department of Informatics and RITMO, University of Oslo, Norway}
	}
	
	\maketitle

\begin{abstract}
	
	
	The structure and performance of neural networks are intimately connected, and by use of evolutionary algorithms, neural network structures optimally adapted to a given task can be explored. Guiding such neuroevolution with additional objectives related to network structure has been shown to improve performance in some cases, especially when modular neural networks are beneficial. However, apart from objectives aiming to make networks more modular, such structural objectives have not been widely explored. We propose two new structural objectives and test their ability to guide evolving neural networks on two problems which can benefit from decomposition into subtasks. The first structural objective guides evolution to align neural networks with a user-recommended decomposition pattern. Intuitively, this should be a powerful guiding target for problems where human users can easily identify a structure. The second structural objective guides evolution towards a population with a high diversity in decomposition patterns. This results in exploration of many different ways to decompose a problem, allowing evolution to find good decompositions faster. Tests on our target problems reveal that both methods perform well on a problem with a very clear and decomposable structure. However, on a problem where the optimal decomposition is less obvious, the structural diversity objective is found to outcompete other structural objectives -- and this technique can even increase performance on problems without any decomposable structure at all.
	
\end{abstract}


\section{Introduction}


The structure and performance of neural networks are closely related, yet the most common technique for training neural networks does not allow structures to change: Only the \emph{weights} of existing connections are modified~\citep{LeCun2015}. The field of \emph{neuroevolution}, where neural networks are optimized with evolutionary algorithms, offers an alternative where both connections and structures can be modified~\citep{Xin1999}. However, there have been only a limited number of studies of how the structure and performance of evolving neural networks are related, and fewer still on the potential for \emph{objectives} related to network structure to guide the evolutionary search. One structural feature which has gained some attention, and been shown to guide evolution when applied as an objective, is \emph{modularity}.

Modularity in evolving neural networks has been demonstrated to improve performance on complex tasks, as it allows problem decomposition, hierarchical knowledge structures and multimodal behavior. There is therefore a growing interest in techniques for increasing the functional modularity of evolving neural networks~\citep{Clune2013, mengistuHierarchy,Schrum2016a, Velez2017}. Most techniques for increasing modularity in neuroevolution belong to one of two extremes.

On one extreme are techniques that explicitly form separate modules for solving separate parts of a problem~\citep{Togelius2004, Cardamone2009, Schrum2016a}. Common to these methods are that there is a clear task division, where it is always known which module solves which subproblem. Examples include layered evolution~\citep{Togelius2004} and multitask networks~\citep{Schrum2012}. We refer to such techniques as \emph{explicit} modularity guidance.

On the other extreme are methods that encourage \emph{general} modularity, where it is not always obvious which module solves which subproblem, or where a very clear modularity may not emerge at all. Examples are methods that evolve modularity by imposing connection costs during evolution~\citep{Clune2013} and methods where the genotype-phenotype mapping tends to lead to high levels of modularity~\citep{Mouret2008, Verbancsics2011, Huizinga2014, gruau1994automatic}. These methods tend to give evolution more freedom to explore different network topologies, at the cost of guiding evolution less towards promising solutions. We refer to these techniques as \emph{implicit} modularity guidance.

\begin{figure}
	\centering
	\begin{subfigure}[b]{0.49\textwidth}
		\includegraphics[width=\textwidth]{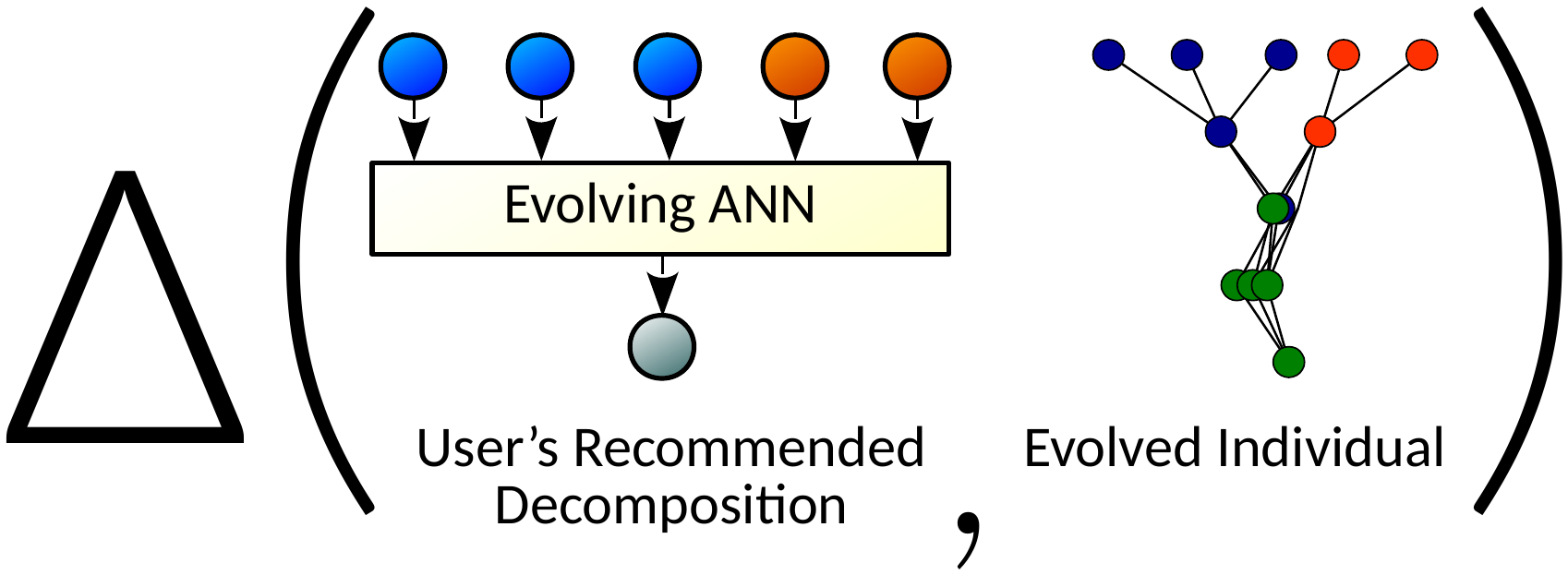}
		\caption{$\Delta_{decomp}(u, i)$: the distance between a user-recommended decomposition $u$, and the evolved modularity in individual $i$.}
		\label{fig:delta_structure}
	\end{subfigure}
	\hfill
	\begin{subfigure}[b]{0.49\textwidth}
		\includegraphics[width=\textwidth]{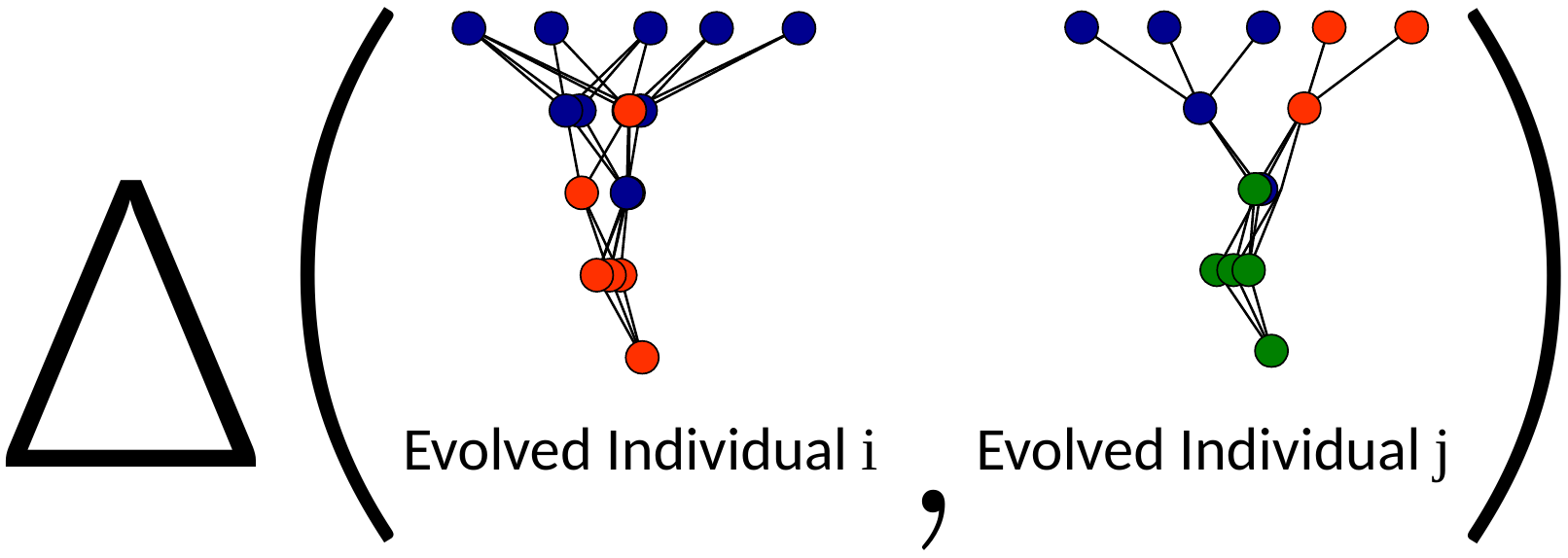}
		\caption{$\Delta_{decomp}(i, j)$: the distance between the modular decompositions in evolved individuals $i$ and $j$.}
		\label{fig:delta_structure_ind_ind}
	\end{subfigure}
	\caption{We propose a distance measure, $\Delta_{decomp}$, with which we compare the modular structure of evolving neural networks to a) A user-defined problem decomposition, and b) The structure of other individuals in the population.}
	\label{fig:concept}
\end{figure}

In this paper, we suggest two new ways to guide evolution towards promising modular decompositions, both relying on a new \emph{distance measure} quantifying the difference between the modular decompositions in two neural networks (Figure~\ref{fig:concept}). The first, \emph{User-defined Modularity} attempts to achieve both the free-form evolution and problem decomposition from the \emph{implicit} modularity methods, and the ability to guide evolution with the user's knowledge of problem decomposition offered by \emph{explicit} modularity guidance. It does this by adding a recommended problem decomposition as an evolutionary objective, guiding the search -- but without \emph{constraining} the search to this specific decomposition pattern.


The second technique we propose takes the opposite approach: Rather than guiding evolution towards a single modular decomposition, it guides evolution towards as many different decompositions as possible, by adding \emph{modular diversity} as an objective.

\begin{figure}
\centering
\includegraphics[width=\textwidth]{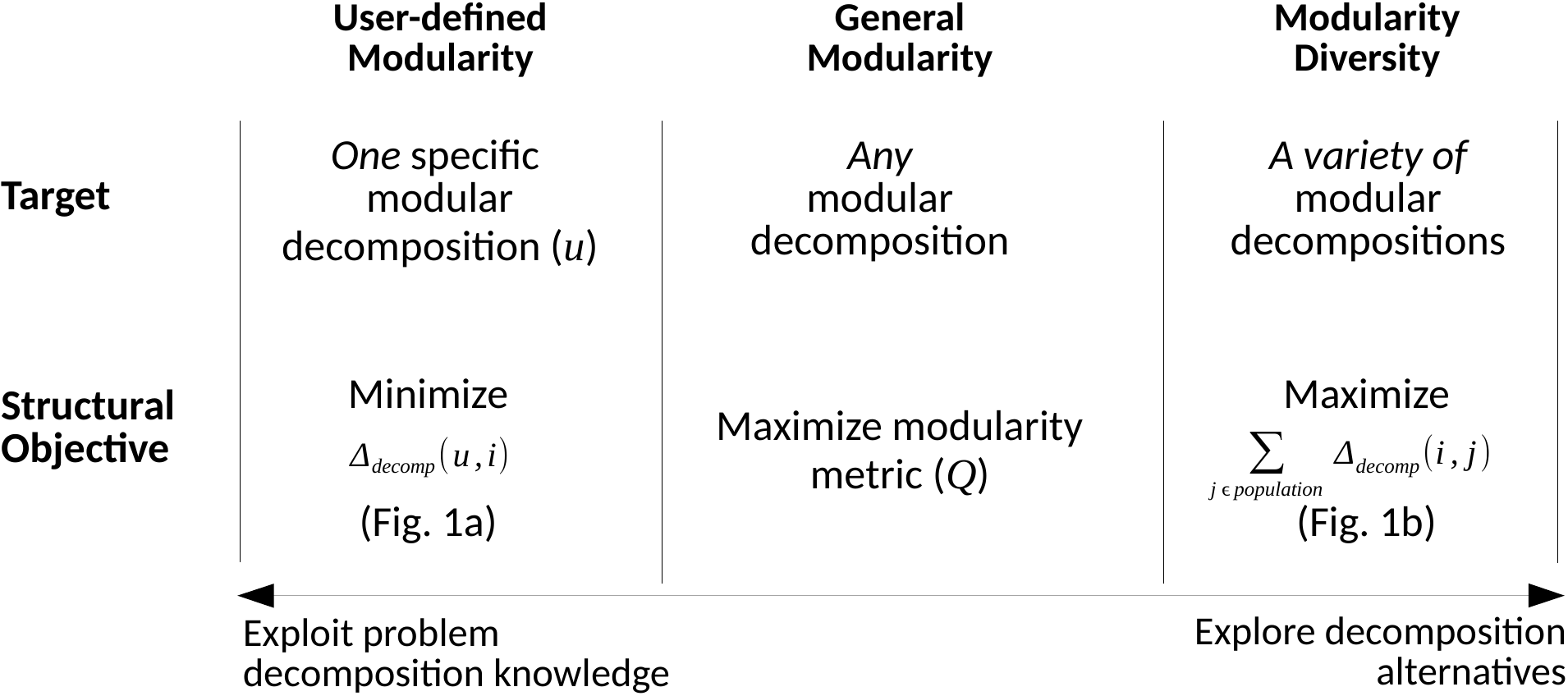}
	\caption{We study the role of structural objectives in guiding evolving neural networks. The applied structural objectives vary in how specific their target structures are. At one extreme, we search for a \emph{single specific modular decomposition}. At the other, we attempt to discover \emph{all useful problem decompositions} by encouraging diversity. In between these extremes, we also test the search for \emph{general modularity}, where modularity itself is the objective, without preference for any specific structure.}
	\label{fig:concept_explore_exploit}
\end{figure}

We compare neuroevolution guided by user-defined modularity, general modularity, and modular diversity on two problems where finding the best modular decomposition can benefit evolution (Figure~\ref{fig:concept_explore_exploit}). Our results indicate that searching for one specific decomposition can work well for problems with a very clear modular structure, but fails when tasks have a less clear structure. However, Modularity Diversity is demonstrated to be a good guide for evolution for both kinds of problems -- even increasing performance on a problem without any modular structure at all.


\section{Related Work}

\subsection{Evolution of neural modularity}

Modularity is here understood as the tendency for a network to have multiple densely connected clusters, each with only a limited connection to other clusters~\citep{Clune2013}. Such modularity is an important organizing principle in many biological networks, including the neural networks that make up the brains of humans and animals~\citep{alon2006introduction,mountcastle1997columnar}. Understanding why modularity evolved in such networks has therefore been a focus of much research, resulting in several different hypotheses on which factors promote the evolution of modularity.

A leading hypothesis has been that modularity evolves when the overall evolutionary goal changes rapidly, but subgoals remain fixed~\citep{Kashtan2005}. Such conditions may have been present in biological evolution in environments that change, but require different combinations of some basic skills or functions. Several other hypotheses have been suggested and demonstrated to lead to the evolution of modularity in simulation, including modularity emerging as a way to reduce interference between different patterns of network activity ~\citep{espinosa2010specialization}, modularity emerging due to a noisy genotype-phenotype mapping~\citep{Hoverstad2011} and modularity emerging due to the costs of building and maintaining neural connections~\citep{Clune2013}.


Unlike the work introduced above, our structurally guiding objectives are not meant to offer a biologically plausible explanation for how modularity evolves in neural networks. Rather, they represent tools which can be used to study the relationship between structure and performance in evolving neural networks, as well as achieving better performing networks by aligning with the problem structure or exploring a wider range of modular decompositions.

\subsubsection{Techniques for leveraging modularity in neuroevolution}

Several researchers have studied how modularity can increase the performance of evolving neural networks. Particularly on complex, decomposable problems, modularity has been demonstrated to allow evolution to find good solutions faster. Evolving modular neural networks can be done in a variety of ways, ranging from explicitly decomposing the problem using domain knowledge, to gently guiding evolution towards modular structures. 


An example of explicit decomposition is layered evolution~\citep{Togelius2004}, inspired by the subsumption architecture from behavior-based robotics~\citep{Brooks1986a}. Layered evolution evolves neural networks incrementally and modularly, beginning with networks for low-level behaviors, and evolving more complex behaviors on top of them as they stabilize. User-knowledge is required for defining the sub-behaviors each module should learn, but the selection among modules during execution is optimized by evolution.

In a series of experiments, Schrum and Miikkulainen~\citep{Schrum2012, Schrum2014, Schrum2016a, Schrum2016b} studied the role of modular neural networks in game scenarios which require multiple skills. In these experiments, separate \emph{output modules} allowed single neural networks to show multiple different behaviors. The researchers studied different ways to perform this modular decomposition, with different degrees of hand design -- ranging from manually specifying which module to use in which situation to giving evolution full control both over the number of different modules and how they are used.

Schrum and Miikkulainen also discuss the importance of task divisions at the input level, highlighting two different ways one can influence neural network architectures~\citep{Schrum2016b}. The first method, called ``split sensors'', is to have separate inputs for sensory information that needs to be processed differently (e.g. different inputs for poisonous and edible objects), thereby biasing learning towards one particular task division. An alternative to this are ``conflict sensors'', which means that these neural-network inputs carry information about multiple different types of events (e.g. a single input that signals both poisonous and edible). The latter make learning harder, but it is important to be able to learn with conflict sensors, since split sensors are not available in all domains.~\cite{Schrum2016b} go on to demonstrate that evolving modular neural networks is a way to learn multimodal behaviors with conflict sensors. We test our methods on a task with a very explicit task decomposition at the input level, similar to ``split sensors'', and on tasks with much less obvious mappings from network inputs/outputs to modular decompositions -- studying the effect of structural objectives also on tasks closer to real-world scenarios, where we are not always sure to which module a neuron belongs.


An alternative to the methods above is to increase modularity in neural networks by letting it emerge without any explicit human design. One such technique adds as an evolutionary objective the reduction of connection costs~\citep{Clune2013} -- resulting in evolved neural networks with increased modularity. This technique has also been found to improve performance on modularity-demanding tasks, such as tasks requiring the learning and retention of multiple skills~\citep{Ellefsen2015}, tasks with multiple subproblems~\citep{Huizinga2014} and tasks with hierarchical structure~\citep{Huizinga2014,mengistuHierarchy}. A related idea is to add \emph{network modularity} itself as an objective. This has also been demonstrated to increase modularity, and in some cases performance, of evolving neural networks~\citep{Huizinga2016}.

A final way to increase modularity in evolving neural networks is to apply genotype-phenotype mappings with modularity-inducing properties~\citep{Mouret2008, Verbancsics2011, Huizinga2014, gruau1994automatic}. An example is applying a developmental process as mapping, which can produce modularity by recursive repetition of developmental rules~\citep{gruau1994automatic}. Similarly to the addition of an objective guiding evolution towards modularity, these techniques encourage modularity in general, but do typically not apply any problem-specific domain knowledge. 


The most explicit methods for evolving modular networks have the advantage of having a clear task division, where it is always clear which module is responsible for each action. On the other hand, the techniques producing modularity by applying guiding objectives or modularity-inducing genotype-phenotype mappings, may give evolution more freedom to explore unconventional modular decompositions, with the disadvantage that these decompositions may be difficult to interpret, and that we do not exploit the user's knowledge about the problem structure.


\subsection{Encouraging diversity in evolving neural networks}

Maintaining diversity in evolutionary algorithms is a commonly used technique to encourage exploration and avoid convergence to local optima~\citep{A.E.Eiben2003a}. Generally, diversity may be encouraged at the \emph{genotype} or \emph{phenotype} level, both requiring an application-specific distance-measure for individuals. An example is \emph{fitness sharing}~\citep{Goldberg1987}, wherein similar individuals \emph{share their fitness value}, resulting in a lower selection pressure on solutions that are very different from others. The popular neuroevolution-algorithm NEAT~\citep{Stanley2002} applies this technique to evolving neural networks, by imposing fitness sharing according to the number of identical genes (which in turn indicate identical nodes or connections). More recently, \emph{multiobjective evolution} has emerged as a way to add diversity as a separate objective, to be optimized together with performance~\citep{Mouret2009a}. 

A key challenge in encouraging diversity in evolutionary algorithms is finding an appropriate way to measure the distance between two individuals. In addition to uncovering \emph{interesting differences} between two individuals, this measure should be efficient to calculate: All individuals in the population need to be compared to all others, resulting in $N^2$ calculations of this value for each generation. In general, computing the distance between graphs (an example of which is neural networks) is NP-hard, ruling out a complete structural distance calculation~\citep{Mouret2009a}. Applying \emph{approximate} structural distance as a guiding objective has been tested as a way to encourage structural diversity in a population of evolving neural networks~\citep{Mouret2009a} -- however, this was \emph{not} found to improve the evolutionary search.

Since structural differences are difficult to calculate, and may not necessarily lead to interesting differences in the functionality of neural networks, a more common technique in evolving neural networks is to apply \emph{behavioral diversity} as an objective~\citep{Mouret2012, Risi2009}. Behavioral diversity techniques rely on quantifying the difference in how evolved individuals \emph{actually behave}. For instance, in a robot navigation task, this could include some information about where the evolved robot tends to navigate to. Encouraging behavioral diversity has been demonstrated to substantially improve the performance of evolutionary algorithms on a variety of different tasks~\citep{Mouret2012} -- and to outperform structural diversification~\citep{Mouret2009a}.

Unlike previous techniques, our structural diversity measure encompasses the idea that the interesting differences between evolving networks lie in their higher-level modular structures, and not in the exact patterns of connectivity. This higher level of abstraction in measuring structural diversity has the additional benefit that high-level differences are faster to compute. We demonstrate that evolution guided by our structural diversity measure performs similarly to evolution guided by behavioral diversity -- and that structural diversity may even lead to faster convergence. A further advantage of structural diversity is that the calculation is independent of the problem -- whereas behavioral diversity techniques typically need some adaptation to a given task~\citep{Mouret2012}.


\section{Targeted Problems}
\label{sec:target_problems}

We apply structurally guided neuroevolution to two problems which are expected to give different insights into the role of structural objectives. The first, the \emph{retina} problem, has a very clear structure and can clearly benefit from one specific decomposition pattern. The second problem, a \emph{robot locomotion problem}, also has a modular structure, but it is not obvious which modular decomposition would work best, or if such a decomposition is necessary.

\subsection{The Retina problem}

\begin{figure}
	\centering
	\begin{subfigure}[b]{0.35\textwidth}
		\includegraphics[width=\textwidth]{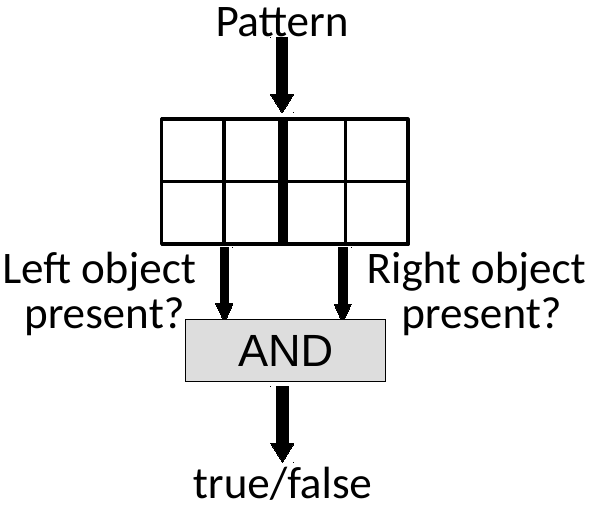}
		\caption{The retina problem.}
		\label{fig:retina_concept}
	\end{subfigure}
	\hfill
	\begin{subfigure}[b]{0.64\textwidth}
		\includegraphics[width=\textwidth]{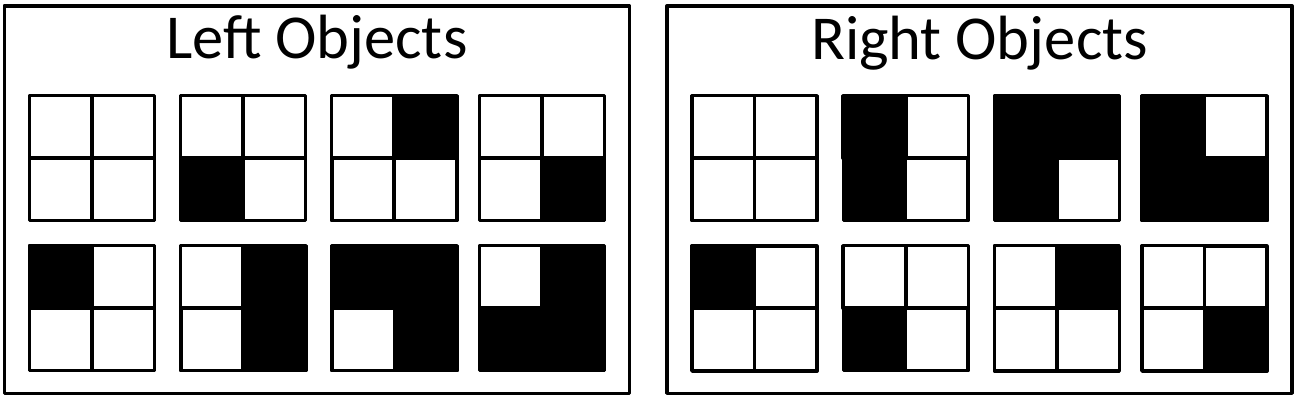}
		\caption{Target patterns for each half of the retina.}
		\label{fig:retina_objects}
	\end{subfigure}
	\caption{In the retina problem, binary patterns are to be classified as true or false, and the input structure allows the problem to be modularly decomposed into two sub-problems. Figures adapted from~\citep{Clune2013}.}
	\label{fig:retina_problem}
\end{figure}

The retina problem (Figure~\ref{fig:retina_concept}) is a pattern-recognition 
task which has been the focus of several previous studies on the evolution of 
modular neural network structures~\citep{Kashtan2005, clune2010investigating, 
Hoverstad2011, Clune2013, Huizinga2014}. In this task an 8-bit input is to be 
classified as 1 or 0. The task is modular, because the input patterns have two 
independent parts (left and right), both of which should contain one of several 
target patterns for the classification to be a 1. We can think of this as 
abstracting the left and right half of a retina, seeing independent parts of 
the visual scene.

Interestingly, even though the retina problem has a modular structure, and evolution can benefit from dividing it into separate parts, evolving neural networks for this problem tends to produce non-modular solutions~\citep{Kashtan2005}. Variants of this problem have therefore been used to gain a better understanding of the environmental pressures that \emph{encourage} the evolution of modularity~\citep{Kashtan2005, Hoverstad2011, Clune2013}.



\subsubsection{Problem setup}
\label{sec:methods_retina}
Our setup of the retina problem follows the ``left AND right'' setup used in previous studies~\citep{Kashtan2005, Clune2013}. The left and right half of the retina both consist of 4 inputs, yielding a total of 16 potential binary patterns on each half of the retina. 8 of these are classified as target patterns (Figure~\ref{fig:retina_objects}).

The task of the evolving networks is to give a positive output whenever the left half of the retina matches one of the left input patterns, \emph{and} the right half of the retina matches one of the right input patterns. Otherwise, it should output a negative number. Each evolving network is presented with all 256 possible input patterns, and the number of errors (wrong answers on the output) are counted. The fitness calculation is 

\begin{equation}
fitness = 1 - \frac{errorCount}{256}
\end{equation}

where $errorCount$ is the number of patterns that were wrongly classified. The fitness score is thus between 0 and 1, where 1 indicates all inputs being correctly classified, and 0 indicating all patterns receiving the wrong classification.

To explore the relationship between problem structure and structural objectives, we also test the techniques on a \emph{nonmodular} version of the retina problem. In this experiment, patterns are distributed randomly across all inputs, eliminating any decomposable problem structure. To keep the problem difficulty similar to the modular retina experiment, we define the same number of target patterns for the nonmodular retina. That is, 64 of the 256 patterns are randomly chosen to be targets.

\subsubsection{Neural network details}

The neural network setup replicates recent work on evolving modular neural networks for the retina problem~\citep{Clune2013}. Evolution optimized the connectivity and connection weights of feed-forward neural networks with a layered structure, only allowing connections between neighboring layers. The output of each neuron, $y_j$, was calculated as the following function of its inputs: $y_j = \tanh( \lambda (\sum_{i \in I} w_{ij}y_i + b))$. $I$ is the set of all inputs to node $j$, $b$ is a bias input, and $w_{ij}$ is the weight of the connection between node $i$ and $j$. The $\tanh$ function ensures an output of each neuron in the range [-1, 1], and $\lambda$ determines the slope of the activation function between the limits. Identically to~\citep{Clune2013}, we set $\lambda$ to 20, making the activation function very steep, resembling a step function. The evolving neural networks had 5 layers, with a maximum of 8/4/2 nodes in each hidden layer. Following~\citep{Clune2013}, evolution chose from a discrete set of values for weights and biases (the values {-2, -1, 1, 2} and {-2, -1, 0, 1, 2} respectively).


\subsubsection{Interface towards neural networks}

The neural networks apply the same input/output structure as previous work on this problem~\citep{Kashtan2005, Clune2013}. The eight binary-valued pixels of the retina are sent to 8 separate input neurons, resulting in 4 neurons receiving the ``left-half'' retina stimuli and 4 others receiving the ``right-half'' stimuli. Output from the network is a single number, with positive output values being interpreted as \emph{true}, and negative outputs as \emph{false}. 

The structural distance between neural networks, $\Delta_{decomp}$, is measured on input-neurons, since these mirror the modular structure of the problem. The recommended decomposition pattern reflects the obvious modular decomposition (Figure~\ref{fig:retina_decomposition}).

\def\imagebox#1#2{\vtop to #1{\null\hbox{#2}\vfill}}
\begin{figure}
	\captionsetup[subfigure]{justification=justified,singlelinecheck=false}
	\centering
	\begin{subfigure}[b]{0.42\textwidth}
		\imagebox{5.2cm}{\includegraphics[width=\textwidth]{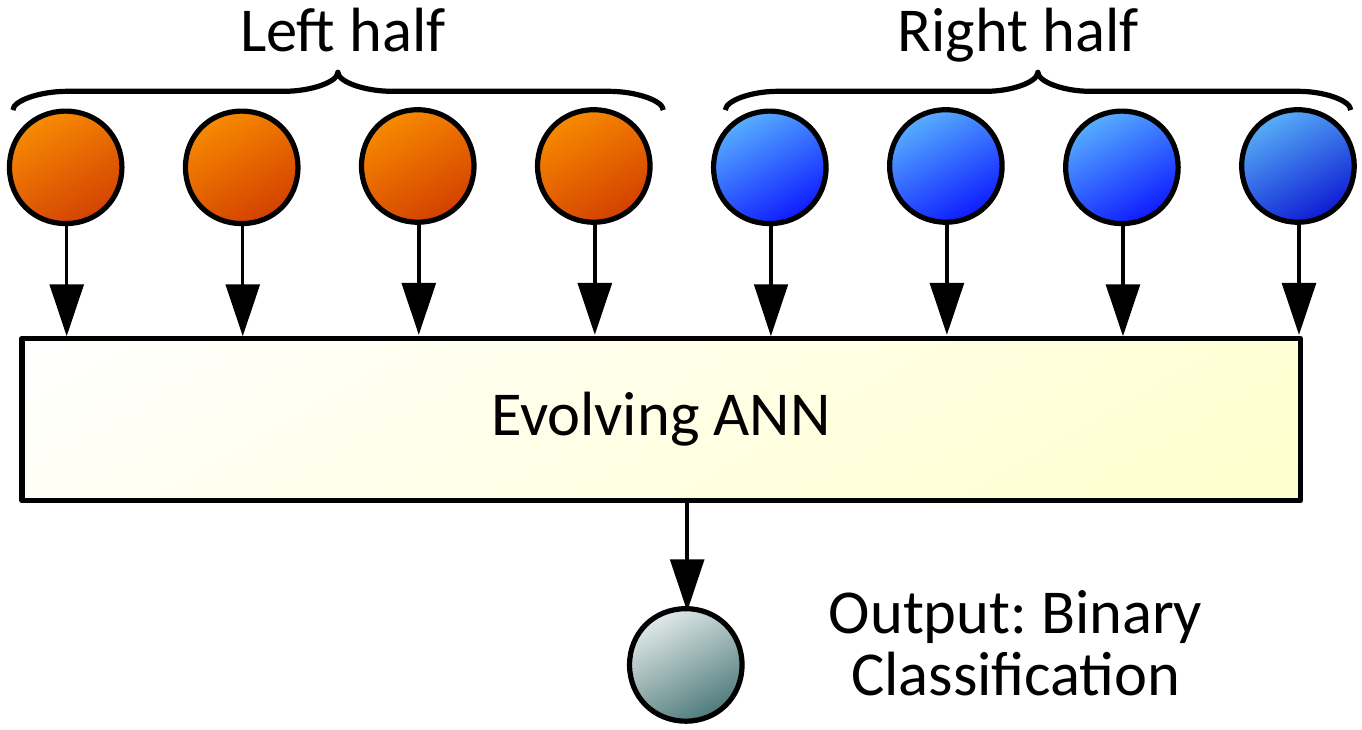}}
		\caption{$M_{rec}$ for the retina problem. Pattern: $[[i_1,i_2,i_3,i_4], [i_5,i_6,i_7,i_8]]$}
		\label{fig:retina_decomposition}
	\end{subfigure}
	~
	\begin{subfigure}[b]{0.55\textwidth}
		\imagebox{6.5cm}{\includegraphics[width=\textwidth]{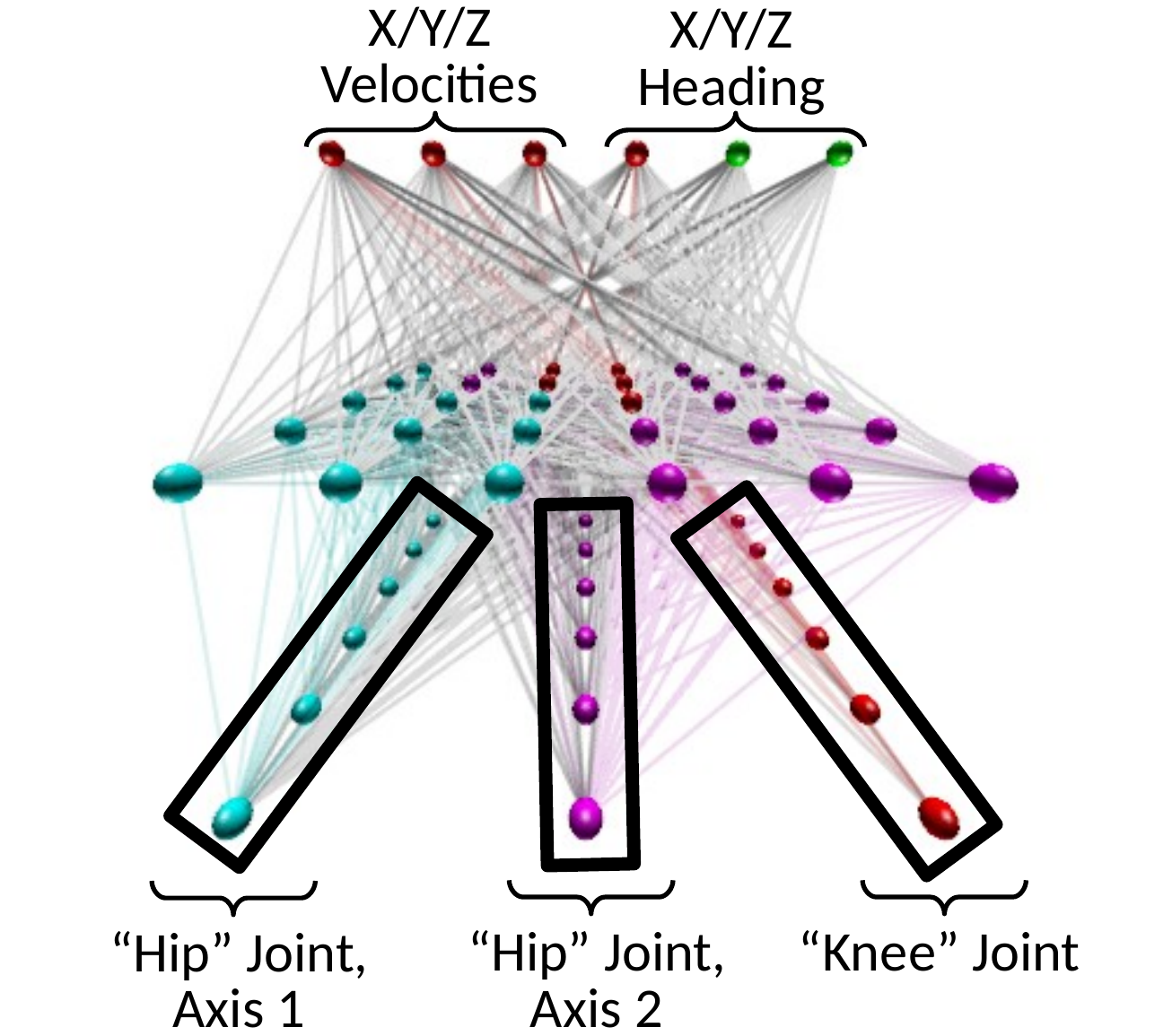}}
		\caption{$M_{rec}$ for the robot locomotion problem. Pattern: $[[o_1-o_6],[o_7-o_{12}], [o_{13}-o_{18}]]$}
		\label{fig:spider_decomposition}
	\end{subfigure}
	\caption{The recommended decomposition patterns $M_{rec}$ for both tasks. For the retina problem, we recommend a decomposition on the ANN input side of the left half and right half of the retina. For the robot locomotion problem, we recommend a decomposition on the ANN output side that groups together the neurons controlling the same axis of motion.}
	\label{fig:applied_decompositions}
\end{figure}

\subsection{Robot Locomotion}

Robot locomotion is a problem that has received significant attention in studies of neuroevolution, including recent experiments on deep neuroevolution \citep{Conti2017a}, and studies on the relationship between structure and performance in evolving neural networks \citep{Huizinga2016}. We test our proposed structural objectives on the robot-locomotion task from \citep{Huizinga2016} to measure their effect on a practical problem, which, unlike the retina problem, does not have a very clear mapping between the structure of the problem and the evolving neural networks.

\subsubsection{Problem setup}

Our setup of the robot locomotion problem follows the ``forward task'' from \citep{Huizinga2016}. In this task, a simulated six-legged robot has to move the center of its body as far as possible in the positive $x$ direction of the environment within $400$ simulator time-steps. With $x_r$ representing the $x$ component of the center of the robot body, fitness is calculated as:

\begin{equation}
fitness = \frac{x_r}{12.5}
\end{equation}

Where $12.5$ is a normalizing constant that was estimated in~\cite{Huizinga2016} by rounding the maximum distance traveled in preliminary experiments.

\subsubsection{Neural network details}

The network is a Continuous-Time Recurrent Neural Network (CTRNN) (\cite{beer1992evolving}, see Supplementary Material Table 4 for the equations), with its parameters specified by the HyperNEAT encoding \citep{stanley2009hypercube}. In the HyperNEAT encoding, the genotype of the network is a Compositional Pattern Producing Network (CPPN) \citep{stanley2007compositional}, which is effectively a neural network that takes as input the coordinates of two neurons and that outputs the weight of the connection between those neurons. Here, the CPPN is extended with a Link Expression Output \citep{Verbancsics2011}, meaning connections are not expressed at all if the value of this output is smaller than zero, and it implements the multi-spatial substrate method (\cite{pugh2013evolving}, see Supplementary Material Figure 1 for an explanation), which is recommended for robotics problems with different input and output modalities. The CPPN is further extended with two additional outputs that specify the bias and time constant of each neuron. To encode these neuron-specific parameters, the CPPN is presented with the coordinates of the relevant neuron as its first inputs, while its other inputs are set to zero. Afterwards, the output of the CPPN is scaled to the desired range, with the CPPN weight and bias outputs scaled to $[-2, 2]$ and the time-constant outputs scaled to $[1, 6]$. For details about the aforementioned methods, we refer the reader to the cited papers. CPPNs are initialized as minimal, fully-connected networks without hidden nodes, weights drawn uniformly from $[0, 1]$ and activation functions are drawn uniformly from the available set of sine, identity, Gaussian, and sigmoid.

%
%
%

The spatial coordinates of the neurons in the CTRNN controlling the spider robot are as depicted in Figure~\ref{fig:spider_decomposition}, where the neurons are shown inside a cube with sides of length 2, centered around the origin such that it extends from -1 to 1 in all dimensions. The extreme neurons shown in this picture all lie at the edge of this cube. 

\subsubsection{Interface towards Neural Networks}

\def\imagebox#1#2{\vtop to #1{\null\hbox{#2}\vfill}}
\begin{figure}
	\captionsetup[subfigure]{justification=justified,singlelinecheck=false}
	\centering
	\includegraphics[width=0.2\textwidth]{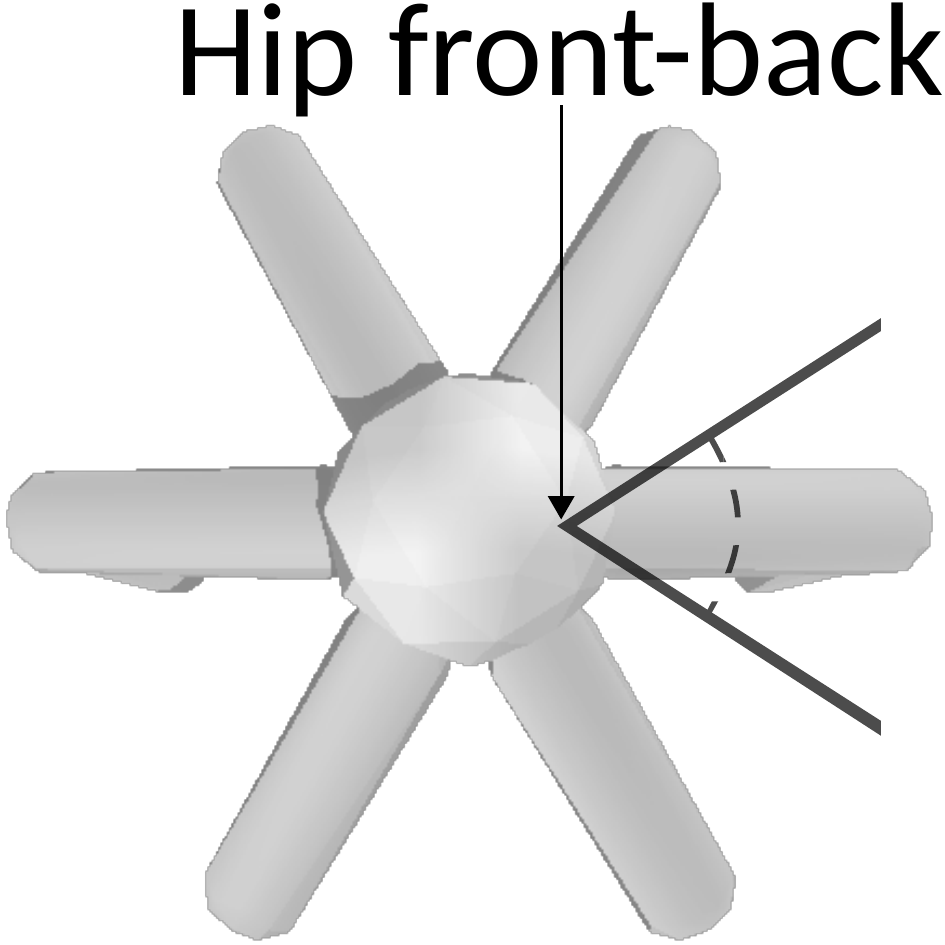}~
	\includegraphics[width=0.2\textwidth]{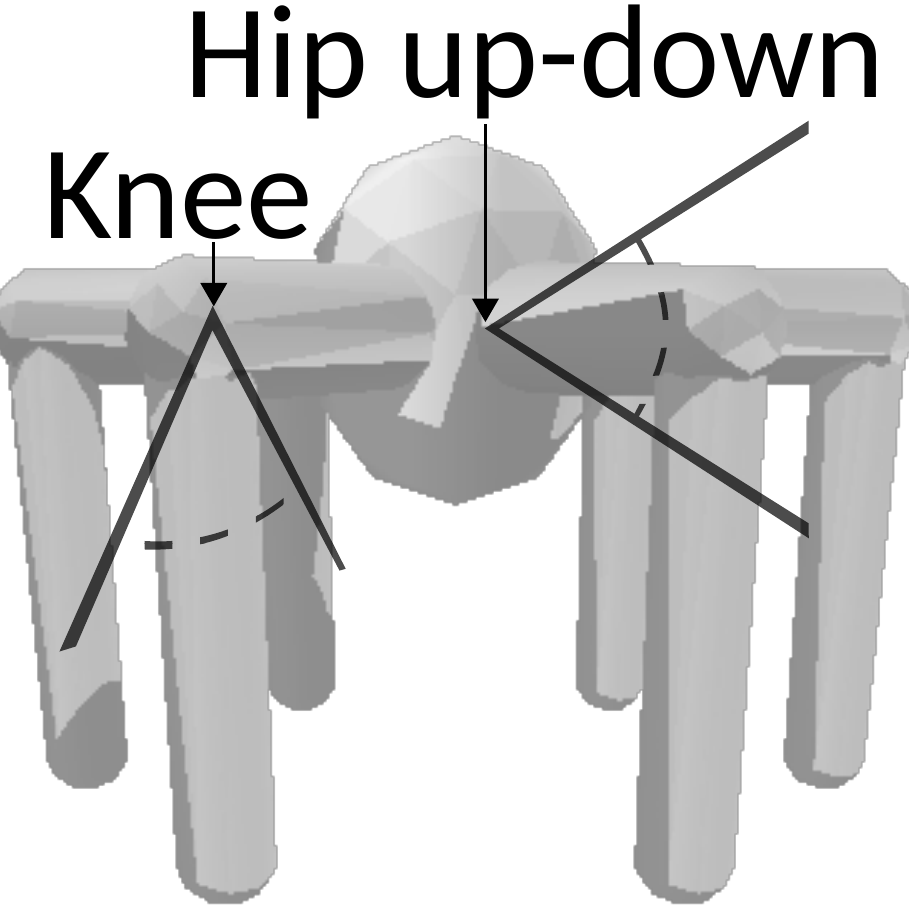}
	\caption{Actuators for the robot locomotion problem. \textit{Figure adapted from~\cite{Huizinga2016}.}}
	\label{fig:robot_actuators}
\end{figure}

As seen in Figure~\ref{fig:spider_decomposition}, the CTRNNs controlling the spider robot have six inputs and 12 outputs. The inputs represent the spider body's velocity along the X, Y and Z- axes, as well as the robot's heading compared to each axis. The heading takes values between +1 and -1, indicating the robot is facing exactly in the direction of the relevant axis, and exactly in the opposite direction, respectively. 

The robot has 6 legs, each with 1 hip joint with 2 degrees of freedom (up-down, front-back) and one knee joint with 1 degree of freedom (Figure~\ref{fig:robot_actuators}). The range of the neural network outputs are rescaled to span the feasible range of their respective actuator and the resulting value is interpreted as the desired angle for that actuator (see Supplementary Material Table 5 for the actuator ranges and velocity calculation).

 This recurrent network can generate rhythmic patterns of activation without 
 any inputs, and initial experiments indicated that many good robot controllers 
 choose to disconnect all inputs to the network. The structural decomposition 
 of inputs is uninformative for such networks, and we therefore use the modular 
 decomposition of \emph{outputs} to calculate the structural distance measure 
 $\Delta_{decomp}$ for this task. It is less obvious how to select a 
 user-recommended decomposition pattern here. One clear structural feature of 
 this task is that each of the robot's six legs has the same three degrees of 
 freedom (Figure~\ref{fig:robot_actuators}), which could potentially benefit 
 from somewhat similar patterns of movement. We therefore recommend a 
 decomposition that divides the output neurons into three groups, one for each 
 of the three degrees of freedom (Figure~\ref{fig:spider_decomposition}).

\section{Methods}

\subsection{Evolving neural networks}

Evolution begins with a population of randomly generated neural networks (for the retina task) or CPPNs (for the robot locomotion task), and works towards better performance by allowing the most fit individuals to have more offspring, and applying random mutations to those offspring. Following previous studies on neuroevolution guided by additional objectives (e.g.~\citep{Clune2013, Ellefsen2017}), we apply the multiobjective optimization algorithm NSGA-II~\citep{Deb2002a}. All individuals have the primary objective of solving the target problem (retina or robot locomotion) as well as possible. Different experimental treatments apply different additional guiding objectives, as outlined in Figure~\ref{fig:concept_explore_exploit}. The experiments were carried out in the Sferes evolutionary algorithm software package~\citep{Mouret2010}. Experimental parameters are given in Supplementary Material Table 1.

Variation in evolving neural networks and CPPNs is created via mutations. For the directly encoded networks in the retina problem, mutations have a small chance of adding connections, removing connections, moving connections, changing the weights of connections and changing bias-inputs to nodes (Supplementary Material Table 2). For the evolving CPPNs in the robot locomotion task, mutations have the potential of modifying connectivity and connection weights, as well as randomly replacing activation functions (Supplementary Material Table 3).

\subsection{User-defined Modularity}
\label{sec:usermod}
Our User-defined Modularity technique allows the user to influence the direction of evolutionary search by defining a modular decomposition that could help solve the target problem. In practice, this is implemented by the user defining a list of lists, where each list corresponds to a module and each element of a list corresponds to a neuron. For instance, the decomposition $[[i_1,i_2], [i_3,i_4]]$ corresponds to a network where input neurons 1 and 2 belong to one module (A), input neurons 3 and 4 belong to a different module (B), and any other input/output neurons are \emph{unspecified}. Unspecified neurons indicate we do not care which module they belong to: They could belong to module A, module B or a different, separate module.


The user-defined modularity pattern, specified in the array-format described above, is given to the multiobjective evolutionary algorithm, which now has the objectives of 1) maximizing \emph{task performance} and 2) maximizing the \emph{degree of match} with the guiding modularity pattern.

%
%
%

\subsection{Quantifying the distance between two modular decompositions}

With the User-defined Modularity technique, it is necessary to evaluate \emph{how well} each evolved network matches the recommended modularity pattern, $M_{rec}$, and with the Modularity-Diversity technique, it is necessary to determine how well the modularity patterns in all \emph{pairs of evolved networks} match. The same distance measure, $\Delta_{decomp}$ (Figure~\ref{fig:concept}), is applied in both cases. In the discussion below, we refer to the modularity pattern in the evolved network we are currently evaluating as $M_{evo}$, and the pattern we are comparing it to as $M_{comp}$. $M_{comp}$ will thus be a user-defined pattern for the User-defined Modularity technique, and the pattern of a \emph{different evolved neural network} for the Modularity-Diversity technique. The calculation of $\Delta_{decomp}$ has two steps: 1) Estimate the modular decomposition of the evolved network(s), and 2) Calculate how well this decomposition matches $M_{comp}$.

\subsubsection{Calculating the modular decomposition of evolved networks}

To evaluate and visualize which are the main modules in an evolved network, we follow a technique applied in previous papers on evolving modular neural networks (e.g.~\citep{Clune2013}). This technique approximates the best modular decomposition of a network, and simultaneously calculates the modularity score of this decomposition. This modularity calculation estimates the network division which maximizes the Q-metric~\citep{Newman2006,Leicht2008}. The Q-metric measures modularity as the difference between the number of connections inside each module and the expected number of such connections for random networks with the same number of edges. In other words, it reflects how ``unexpectedly modular" a given network is. Maximizing Q is an NP-hard problem, and we therefore apply an approximate optimization algorithm to find the most modular division~\citep{Fortunato2010}. More details on this technique can be found in~\citep{Ellefsen2015}.

The result is an estimate of which are the most prominent modules in our evolved neural networks, and the modularity Q-score associated with this modular decomposition. In visualizations (e.g. Figure~\ref{fig:concept}) we color the different discovered modules in different colors.

\subsubsection{Comparing \texorpdfstring{$M_{evo}$}{M-evo} to \texorpdfstring{$M_{comp}$}{M-comp}}
The match between an evolved modular decomposition, $M_{evo}$ and a different decomposition, $M_{comp}$ (either a user recommendation or a different evolved network) is reflected in the metric $\Delta_{decomp}$. Since our evolving neural networks can choose to connect or disconnect internal neurons, we limit this calculation to input and/or output neurons, depending on where we find it most relevant to measure modular decompositions. For the \emph{retina} problem, input neurons mirror the modular structure of the task, whereas for the \emph{robot locomotion} problem, the output neurons have the clearest modular decomposition (and many good solutions do not connect the inputs at all). We therefore measure $\Delta_{decomp}$ on inputs for the former, and on outputs for the latter (Figure~\ref{fig:applied_decompositions}). For simplicity, we discuss measurements on ANN inputs below, but the same calculations apply to measuring $\Delta_{decomp}$ on outputs, or even on internal neurons.

\begin{figure}
	\centering
	\begin{subfigure}[b]{0.35\textwidth}
		\includegraphics[width=\textwidth]{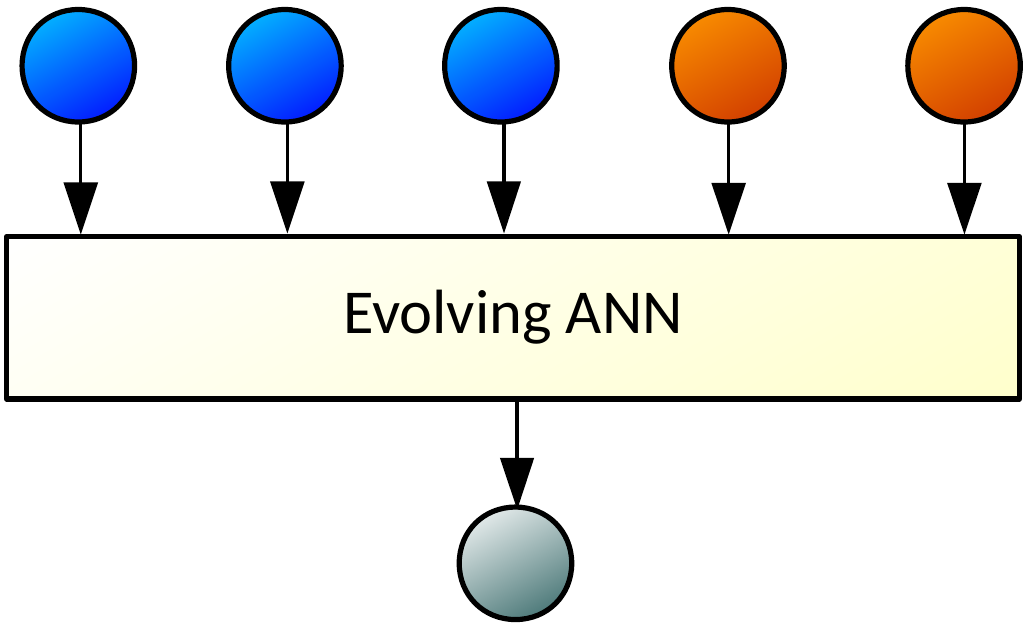}
		\caption{$M_{rec}$: $[[i_1,i_2,i_3],[i_4,i_5]]$}
		\label{fig:m_recommended}
	\end{subfigure}
	\hfill
	\begin{subfigure}[b]{0.29\textwidth}
		\includegraphics[width=\textwidth]{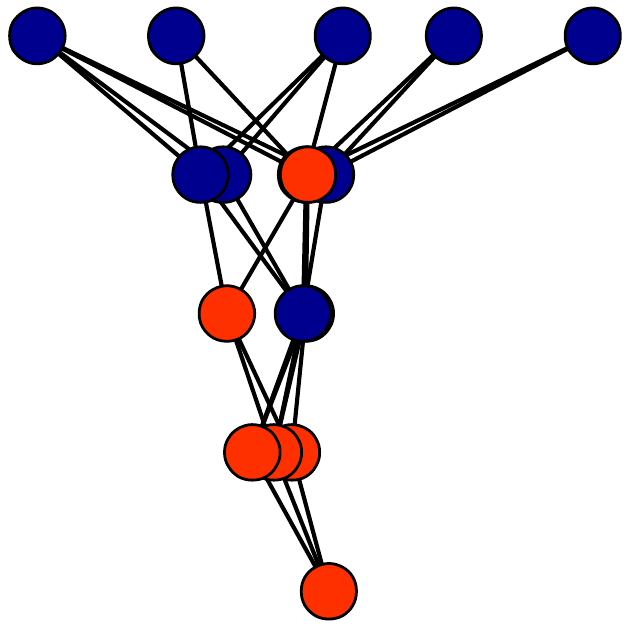}
		\caption{$M_{evo}:$ $[[i_1,i_2,i_3,i_4,i_5]]$}
		\label{fig:uniformity}
	\end{subfigure}
	\hfill
	\begin{subfigure}[b]{0.29\textwidth}
		\includegraphics[width=\textwidth]{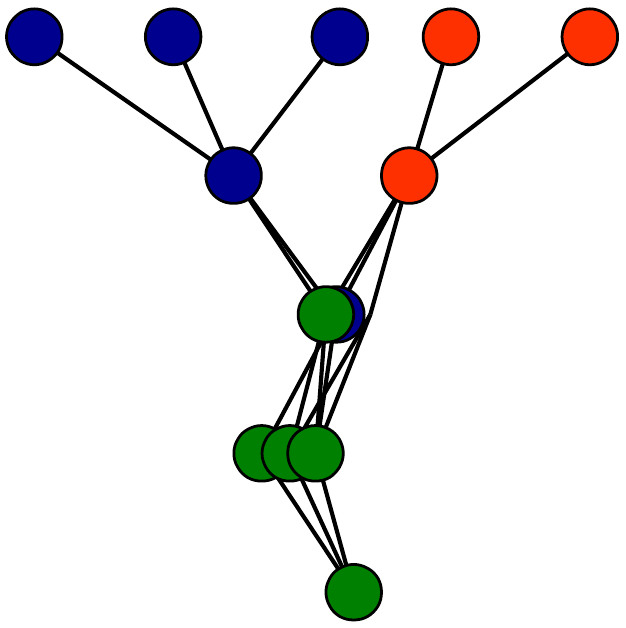}
		\caption{$M_{evo}$: $[[i_1,i_2,i_3],[i_4,i_5]]$}
		\label{fig:uniformity_no_conflicts}
	\end{subfigure}
	\caption{Examples of a) a recommended modular decomposition, b) an evolved network with full \emph{uniformity} but several \emph{conflicts} and c) an evolved network perfectly aligned with $M_{rec}$ (full uniformity and no conflicts).}
	\label{fig:recommended_and_actual}
\end{figure}

When comparing decompositions, we are interested in which neurons belong to the 
same, and which belong to different modules. Other than their constituent 
neurons, modules have no identity -- the color we display to tell modules apart 
has no special meaning. For this reason, we cannot compare two neural networks 
by counting whether their neurons agree on which module they belong to. For 
instance, in Figure~\ref{fig:recommended_and_actual}, it does not matter that 
neuron $i_1$ in $M_{rec}$ and $M_{evo}$ both belong to the ``blue" module. 
However, 
it \emph{does} matter that neurons $i_1$-$i_3$ in both $M_{rec}$ and $M_{evo}$ 
belong to the same module, and that neuron $i_1$ and $i_4$ belong to different 
modules, both in $M_{rec}$ and $M_{evo}$. We therefore need a measure that 
reflects to which degree neurons in $M_{comp}$ and $M_{evo}$ are grouped 
together in the same way.

There are two separate issues that are important when comparing $M_{comp}$ and 
$M_{evo}$. The first is that the neurons belonging to the same module in 
$M_{comp}$ should as far as possible also do so in $M_{evo}$. For instance, for 
the recommended pattern in Figure~\ref{fig:m_recommended}, evolved networks 
will have the lowest $\Delta_{decomp}$ if neurons $i_1$ to $i_3$ belong to the 
same module, and neurons $i_4$ and $i_5$ are also grouped together. We call 
this measure \textbf{uniformity}, as it reflects to what degree neurons that 
were intended to belong to the same module actually do so.

Note that having a high uniformity is not enough for two decompositions to be a 
good match. For instance, if $M_{evo}$ has all neurons belonging to the same 
module, it will score maximally on uniformity no matter how $M_{comp}$ looks: 
All the modules in $M_{comp}$ are 100\% uniform in $M_{evo}$. We therefore need 
to measure also how frequently pairs of neurons in $M_{evo}$ belong to the same 
module, but their counterparts in $M_{comp}$ do not. This is for instance the 
case for neurons $i_3$ and $i_4$ in Figure~\ref{fig:uniformity}: They belong to 
the same module, but were recommended not to do so. We call such a situation a 
\textbf{conflict}. To evaluate how well aligned $M_{evo}$ and $M_{comp}$ are, 
$\Delta_{decomp}$ needs to reflect the degree of \emph{uniformity} inside 
recommended modules, and the degree of \emph{conflict} between them.

We facilitate some explanations below by discussing the ``color'' of modules. As discussed in Section~\ref{sec:usermod}, modular decompositions are lists of lists of neuron IDs. When we discuss modules or neurons of different ``color'', we simply mean that these belong to \emph{different sublists}.

\subsubsection{Calculating uniformity}

Algorithm~\ref{alg:uniformity} calculates the \textbf{uniformity} of two decompositions. The inputs are the evolved and compared modular structures, both presented as lists of lists of neuron-IDs, as seen in Figure~\ref{fig:recommended_and_actual}. The algorithm goes through each module in $M_{comp}$, and  calculates the \emph{uniformity} (to which degree they \emph{also belong to a single module}) of corresponding nodes in $M_{evo}$.

For each module in $M_{comp}$, the method first extracts the IDs of all neurons in that module, $neuron\_ids$. In $M_{evo}$, we then check to which module(s) those neurons belong (which \emph{color} they have in the modular decomposition). The most common color among these neurons is referred to as the $main\_color$ of this module, and counting how many of the neurons in $neuron\_ids$ have that color gives us an indication of the module's uniformity. The uniformity is summed over all modules in $M_{rec}$, and normalized to be in the range [0,1] where higher values indicate more uniformity.

An example of maximum uniformity is shown in Figure~\ref{fig:uniformity}. 
$M_{rec}$ (Figure~\ref{fig:m_recommended}) has 2 modules, consisting of neurons 
$i_1$-$i_3$ and $i_4$-$i_5$ respectively.  The uniformity calculation processes 
these two sequentially. First, it is found that neurons $i_1$-$i_3$ indeed all 
belong to the same module in the evolved network, adding 3 to 
$uniform\_neurons\_counter$. Next, the same is found for neurons $i_4$-$i_5$, 
adding 2 to $uniform\_neurons\_counter$. The final uniformity is therefore 5/5 
= 1. By similar reasoning, the network in 
Figure~\ref{fig:uniformity_no_conflicts} is also found to be fully uniform with 
respect to $M_{rec}$. These two figures illustrate why uniformity alone is not 
a sufficient measure of $\Delta_{decomp}$: Both have a full uniformity with 
$M_{rec}$, but only the evolved network in 
Figure~\ref{fig:uniformity_no_conflicts} has the intended structure.

\begin{algorithm}
\caption{\textproc{$Uniformity$}($M_{evo}, M_{comp}$)}
\begin{algorithmic}[1]
	\State $uniform\_neurons\_counter \gets 0$
	\For{$module \in M_{comp}$}
		\State $neuron\_ids \gets GetNeuronIDs(module)$
		\State $evolved\_neuron\_colors \gets GetColors(M_{evo}, neuron\_ids)$
		\State $main\_color \gets MostCommonColor(evolved\_neuron\_colors)$
		\State $module\_uniformity \gets CountOccurences(main\_color, evolved\_neuron\_colors)$
		\State $uniform\_neurons\_counter \pluseq module\_uniformity$
	\EndFor
	\State \Return $uniform\_neurons\_counter / TotalNumNeurons(M_{comp})$
\end{algorithmic}
\label{alg:uniformity}
\end{algorithm}

\subsubsection{Calculating conflicts}

Algorithm~\ref{alg:conflicts} calculates the number of \textbf{conflicts} between $M_{comp}$ and $M_{evo}$. A conflict exists when neurons from a single module in $M_{evo}$ belong to several different modules in $M_{comp}$. The algorithm goes through all modules in $M_{comp}$, and extracts the colors of the corresponding neurons in $M_{evo}$. The inner loop goes through \emph{all other} modules in $M_{comp}$ to see if any of the same colors can be found on their corresponding neurons in $M_{evo}$. For each such match, we count one conflict. The conflict measure is normalized to lie in the range [0,1] where higher numbers mean fewer conflicts. This is done to make 1 indicate the highest level of agreement for both uniformity and conflicts.

To give an example, in Figure~\ref{fig:m_recommended}, there are two 
recommended modules. Algorithm~\ref{alg:conflicts} starts by assigning the 
first (neurons $i_1$-$i_3$) as $module\_A$ and the second (neurons $i_4$-$i_5$) 
as $module\_B$. In the evolved network in Figure~\ref{fig:uniformity}, the 
colors in $module\_A$ are [blue, blue, blue] and in $module\_B$ [blue, blue]. 
$CountMatches$ goes through all the neurons in $module\_A$, and counts how many 
of the neurons in $module\_B$ match their color. In this case, we have $3*2=6$ 
matches. The normalizing factor $max\_num\_conflicts$ is incremented by the 
maximum number of conflicts between these two modules, which also happens to be 
$3*2=6$. For this network, $max\_num\_conflicts$ is equal to $num\_conflicts$, 
resulting in a conflict measure of 0 (the worst possible). A similar 
calculation on the evolved network in Figure~\ref{fig:uniformity_no_conflicts} 
reveals it has the best possible conflicts-score of 1.

\begin{algorithm}
\caption{\textproc{$Conflicts$}($M_{evo}$, $M_{comp}$)}
\begin{algorithmic}[1]
	\State $num\_conflicts \gets 0$
	\State $max\_num\_conflicts \gets 0$
	\For{$module\_A \in M_{comp}$}	
		\State $module\_A\_neuron\_ids \gets GetNeuronIDs(module\_A)$
		\State $module\_A\_colors \gets GetColors(M_{evo}, module\_A\_neuron\_ids)$
		\State $other\_compared\_modules \gets (M_{comp} - module\_A)$
		\For{$module\_B \in other\_compared\_modules$}
			\State $module\_B\_neuron\_ids \gets GetNeuronIDs(module\_B)$
			\State $module\_B\_colors \gets GetColors(M_{evo}, module\_B\_neuron\_ids)$
			\State $num\_conflicts \pluseq CountMatches(module\_A\_colors, module\_B\_colors)$
			\State $max\_num\_conflicts \pluseq size(module\_A)* size(module\_B)$
		\EndFor			
	\EndFor	
	\State \Return $(max\_num\_conflicts-num\_conflicts)/max\_num\_conflicts$
\end{algorithmic}
\label{alg:conflicts}
\end{algorithm}

\subsubsection{Calculating \texorpdfstring{$\Delta_{decomp}$}{Delta-decomp}}

Finally, the match between two modular decompositions is calculated by taking the average of the level of uniformity and conflict between the two:

\begin{equation}
\begin{aligned}
\Delta_{decomp} & = 1 - \frac{Uniformity(M_{evo}, M_{comp}) + Conflicts(M_{evo}, M_{comp})}{2.0}
\end{aligned}
\end{equation}

$\Delta_{decomp}$ thus ranges from 0 to 1, where 0 indicates a perfect match between the compared decompositions, and 1 indicates the worst possible match.

\subsection{Diversity measurement}
\label{sec:diversity_measurement}

\subsubsection{Behavioral distance}

In one experiment, we compare the use of behavioral and modular diversity in evolving neural networks. A key difference in these two approaches is that behavioral differences typically have to be calculated with problem-specific methods, whereas, using our $\Delta_{decomp}$ metric, modular diversity can be calculated the same way for any problem with neural network phenotypes.

A \emph{generic} behavior distance metric, which has been found to work well for several problems in evolutionary robotics, is the Hamming distance between sensory-motor vectors~\citep{Mouret2012}. The idea in this approach is to store all inputs and outputs of a neural network in a large binary vector (a process which may require some problem-specific adaptation), and calculate the Hamming distance (the number of positions at which the bits are different) between pairs of networks.

Inspired by this, the behavioral descriptors for both our tasks reflect the idea that the behavior of a network is considered different if its response to a particular input is different from the response of the rest of the population. Both diversity measurements are based on representing the history of network outputs as binary vectors, by converting positive outputs to 1 and zero-valued or negative outputs to 0. For the retina task, inputs are always presented to the neural networks in the same order, and simply appending each binary output to a vector generates a description of how the network ``behaves'' as it sees each unique input.

For the robot locomotion problem, behaviors are more complex, since the input to the neural network depends on the previous motions of the robot. To characterize network behaviors here, we use a measure of behavioral diversity similar to the one presented in \citep{Huizinga2016}: We give each network a collection of pre-defined inputs, and measure its response as follows. Setting one of the inputs to 1 and all others to 0, we record the output of the network over 5 time steps, converting it to a binary vector with length equal to the number of outputs. This process is repeated for each input, yielding a behavioral descriptor of length $5*num\_inputs*num\_outputs=5*6*18=540$. Note that even though successful networks for this task sometimes do not connect to the inputs, this method can capture behavioral differences, since the pattern of outputs varies depending on the evolved CTRNN, even without any inputs.

\subsubsection{Measuring diversity against the population}
\label{sec:diverstiy_against_population}
For both the behavioral and modular diversity objective, we follow the recommendation from~\citep{Mouret2012} in calculating the diversity score of an individual as \emph{the average distance} to the rest of the population:

\begin{equation}
\label{eq:behavior_score}
Diversity(x_i) = \frac{1}{N}\sum_{j=0}^{j=N} d(x_i,x_j)
\end{equation}

where $x_i$ is the individual of which we are measuring diversity, $N$ is the population size and $d(x_i,x_j)$ is the distance between individual $i$ and $j$. For the Modularity Diversity, this is equal to $\Delta_{decomp}(x_i,x_j)$, whereas for the behavioral diversity calculation, this is the Hamming distance between the behavior vectors.

\subsection{Experimental treatments}
\label{sec:treatments}

\begin{table*}
\centering
\begin{tabular}{|l|l|}\hline
Treatment & Structural Objective\\ \hline\hline
PA & None\\
UserMod & Maximizing match with user-defined modularity pattern (Figure~\ref{fig:applied_decompositions})\\
Q-Mod & Maximizing modularity as measured with the Q-metric\\
ModDiv & Maximizing diversity of modular decompositions in the population\\
\hline
\end{tabular}
\caption{The different experimental treatments.}
\label{table:treatments}
\end{table*}



Our main experiment compared three different ways of guiding neuroevolution with structural objectives (Table~\ref{table:treatments}). The baseline treatment is ``Performance Alone" (\textbf{PA}), where evolution is guided only by performance on the target problem. In this single-objective case, NSGA-II is an elitist evolutionary algorithm with tournament-based selection. \textbf{UserMod} applies the User-defined Modularity-technique, inserting knowledge about the recommended problem decomposition in the evolutionary search. \textbf{Q-mod} guides evolution towards more modular neural networks, but without applying any problem-specific knowledge. Previous work has shown such general modularity pressure to form more modular~\citep{Huizinga2016} and better performing~\citep{Clune2013} neural networks when applied to modularly decomposable tasks. Finally, \textbf{ModDiv} applies the Modularity-Diversity technique, selecting for networks with different modular decompositions than the rest of the population.


\subsection{Metrics and visualizations}



When calculating the structural modularity of evolved networks, we apply the widely used Q-score~\citep{Newman2006}. In visualizations, we follow~\citep{Clune2013} in first moving nodes to the position that minimizes the length of the neural network, while holding inputs and outputs fixed. This shows structural modularity more clearly, while not changing the functionality or modularity score of the network. Also following~\citep{Clune2013}, in our visualizations, we estimate the most modular split of the network, and color each neuron according to which module it belongs to.

All experimental treatments were repeated 50 times with different stochastic events (that is, using different random seeds). Analyses of evolved networks focus on the best performing network (with regards to the primary objective, and the secondary objective used to break ties) at the end of each trial. All tests of statistical significance apply the Mann-Whitney U test.

\section{Results and Discussion}

To understand the role of the structural objectives outlined in Figure~\ref{fig:concept} in guiding neuroevolution, we measure performance, modularity and diversity in evolving populations guided by each structural objective. We also compare the Modularity Diversity objective to the powerful technique of encouraging \emph{behavioral} diversity. Finally, we test the techniques on a non-modular problem, demonstrating that also problems without any obvious structure can benefit from a structurally diverse population. Figures present medians, bootstrapped 95\% confidence intervals, and markers where there are significant differences between an indicated treatment and the others.

\subsection{Performance}

\subsubsection{The retina problem}
\begin{figure}[h]
	\centering
	\begin{subfigure}[b]{0.49\textwidth}
		\includegraphics[width=\textwidth]{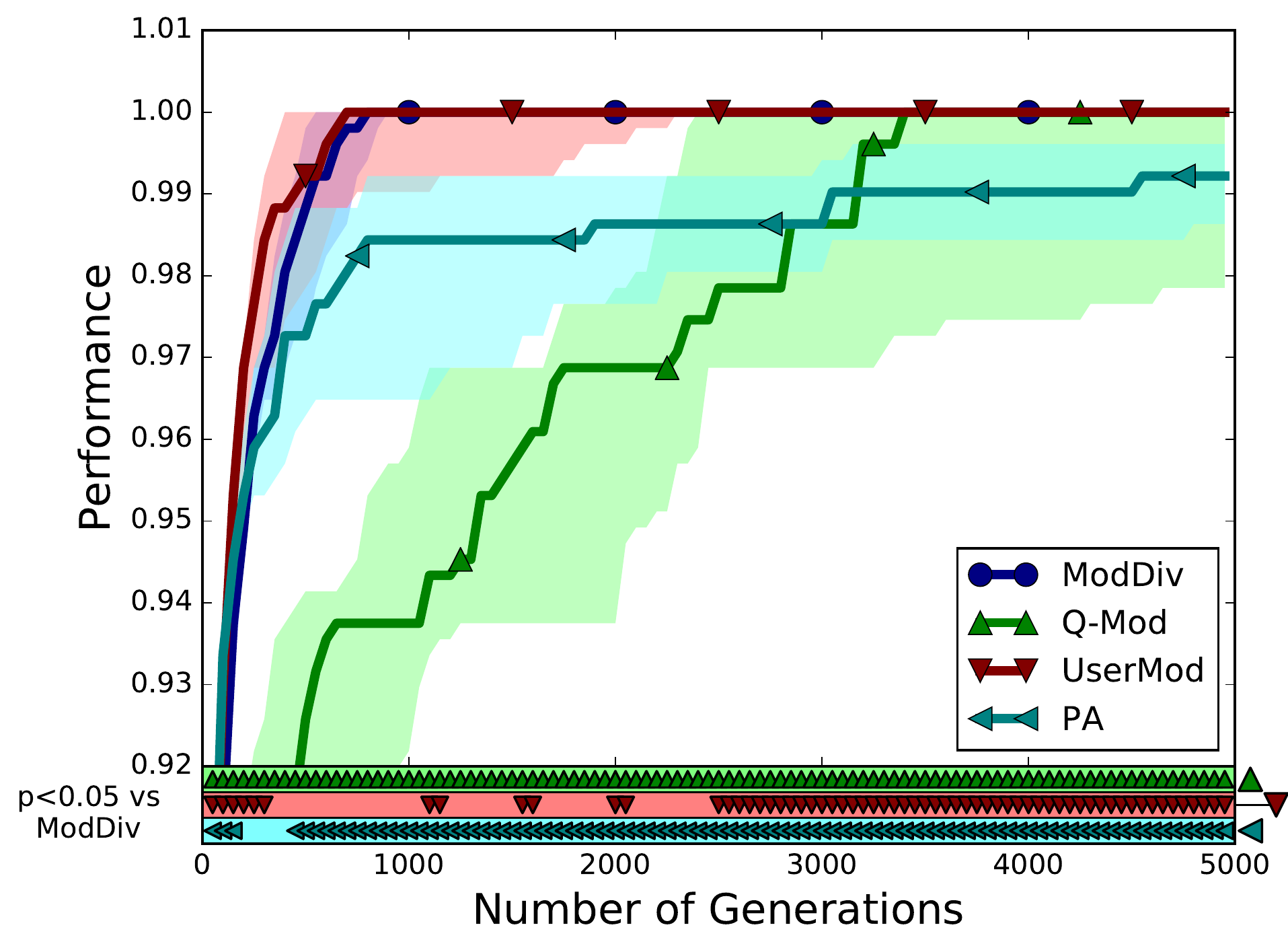}
		\caption{Performance of the best individual}
		\label{fig:performance_retina}
	\end{subfigure}
	\begin{subfigure}[b]{0.49\textwidth}
		\includegraphics[width=\textwidth]{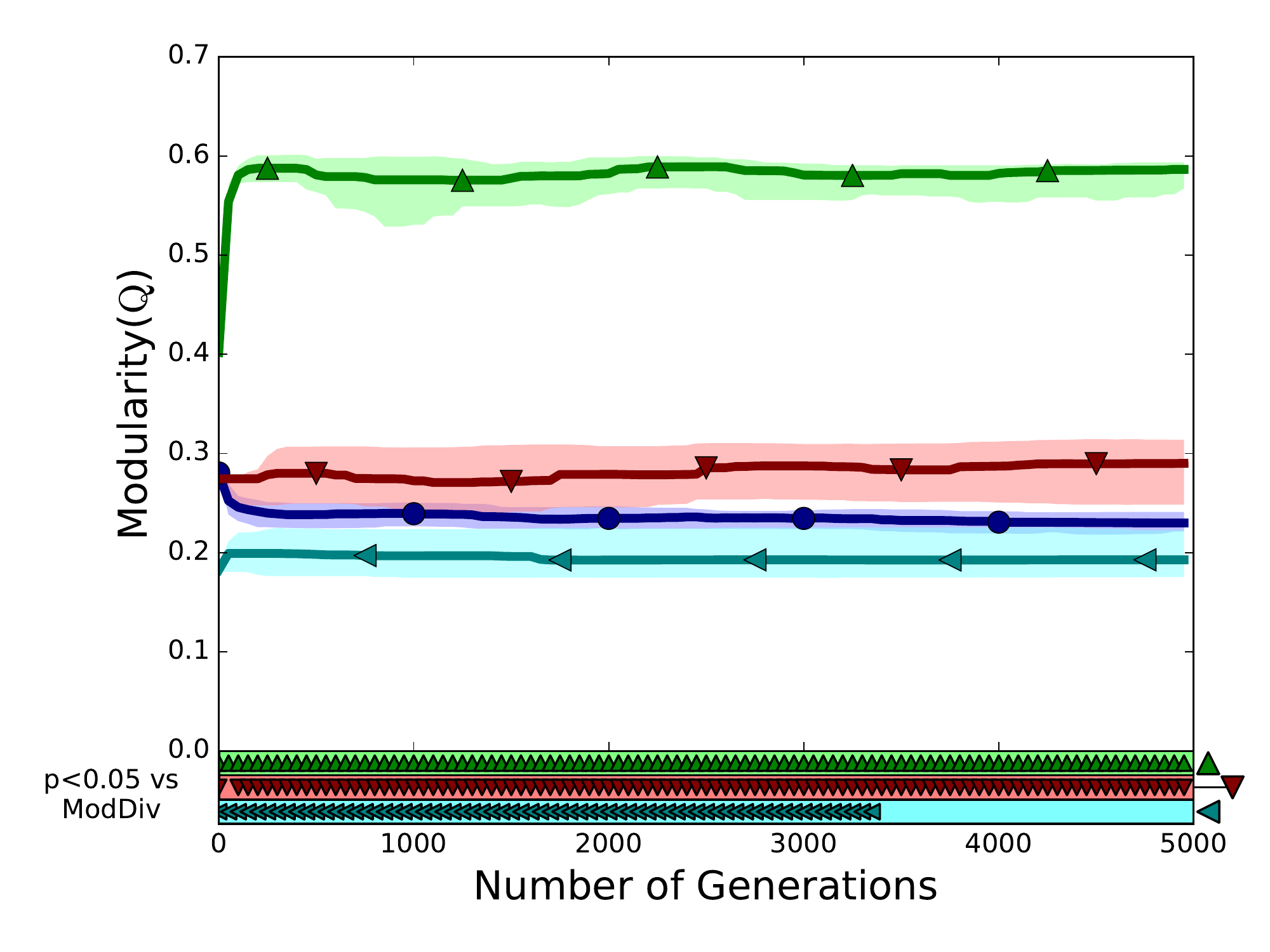}
		\caption{Median population modularity (Q-score)}
		\label{fig:modularity_retina}
	\end{subfigure}
	
	\begin{subfigure}[b]{0.49\textwidth}
		\includegraphics[width=\textwidth]{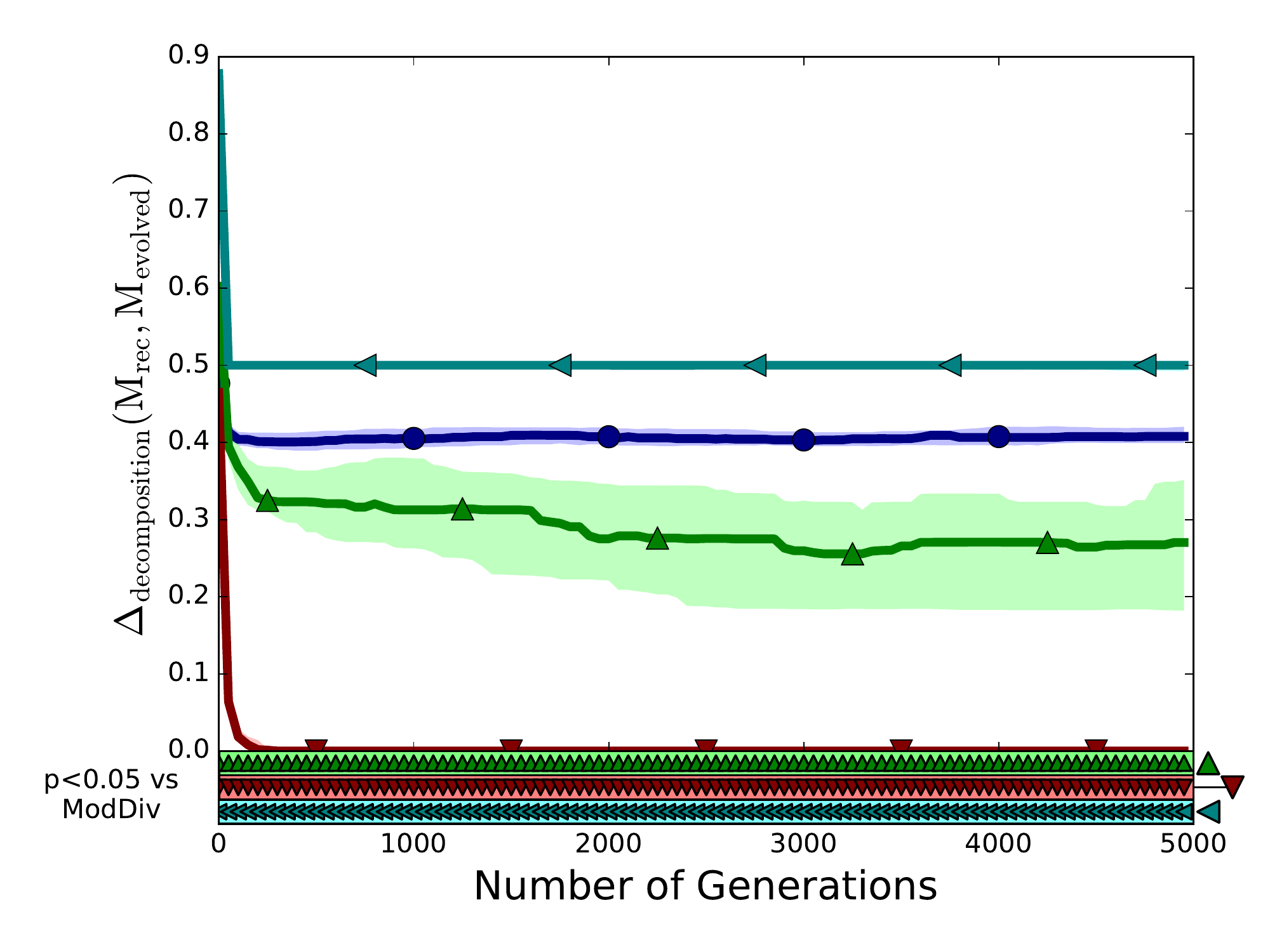}
		\caption{Median population distance from $M_{rec}$}
		\label{fig:usermod_retina}
	\end{subfigure}
	\begin{subfigure}[b]{0.49\textwidth}
		\includegraphics[width=\textwidth]{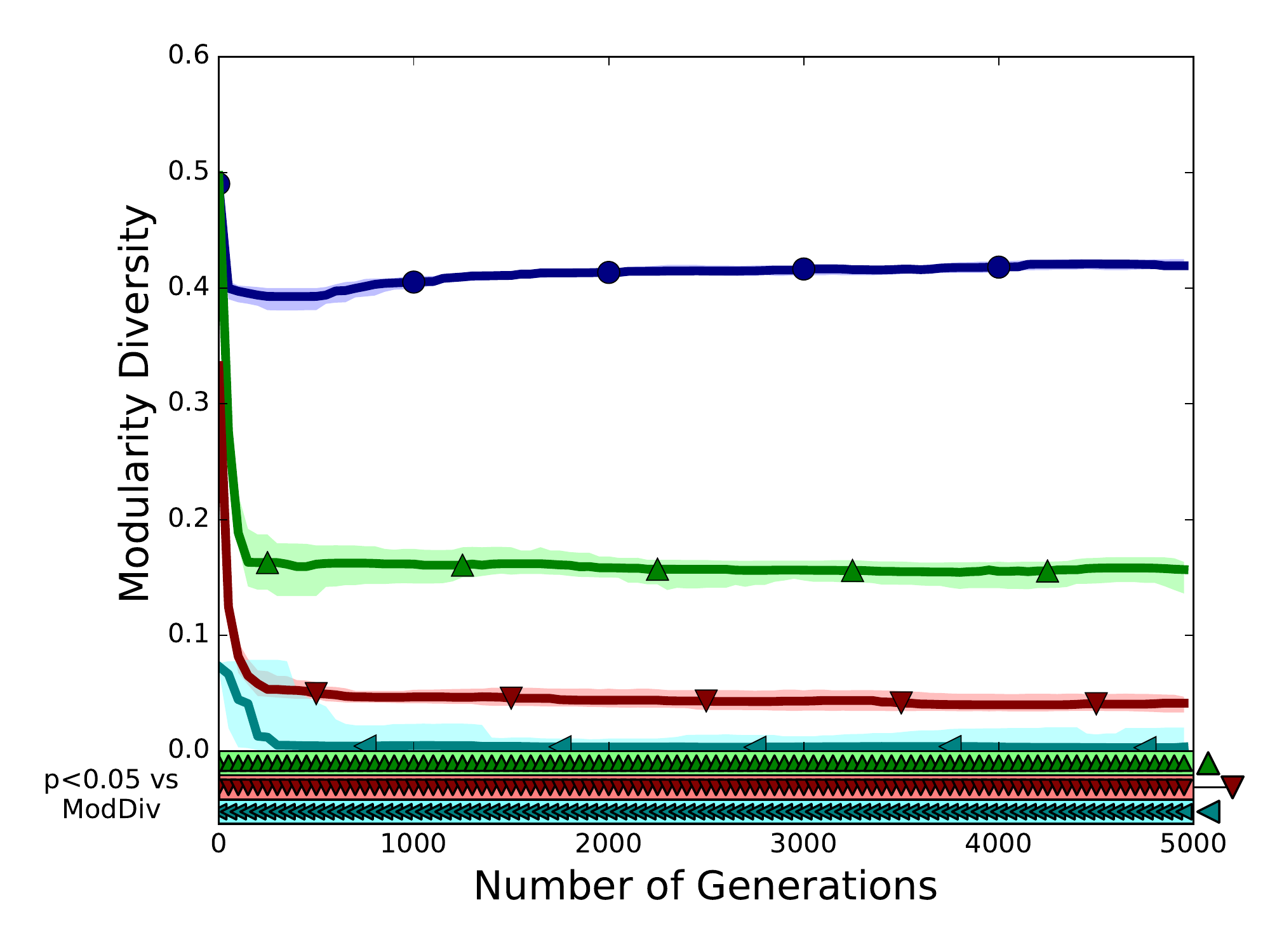}
		\caption{Median population diversity}
		\label{fig:retina_diversity}
	\end{subfigure}
	\caption{The performance and development of different modular decompositions during evolution on the retina-problem.  Perfect performance is reached fastest by the User-defined Modularity and Modularity Diversity treatments.}
	\label{fig:retina_results}
\end{figure}

We compared neuroevolution guided only by the \emph{performance} of evolving networks (PA) to that guided by each of the structural objectives outlined in Table~\ref{table:treatments}. On the clearly decomposable \emph{retina}-problem, the treatments converging fastest are the two opposites of 1) searching for \emph{a diverse set} of modular decompositions (\emph{ModDiv}) and 2) searching for a \emph{single, user-defined} modular decomposition (\emph{UserMod}) (Figure~\ref{fig:performance_retina}).

As expected, the general modularity objective (\emph{Q-Mod}) produces significantly more modular networks than all other treatments (Figure~\ref{fig:modularity_retina}). We also observe that the user-defined structural guidance has a dramatic effect on the problem decomposition in evolving networks: Within a few hundred generations this objective leads the average network to be fully aligned with the user-defined problem decomposition (Figure~\ref{fig:usermod_retina}).

Finally, we measure the average structural diversity in the population. The result (Figure~\ref{fig:retina_diversity}) confirms the ordering we suggested in Figure~\ref{fig:concept_explore_exploit}: User-defined modularity results in the least structural diversity, since it exploits a single modular decomposition. Structural diversity as a guiding objective (\emph{ModDiv}) has the opposite effect: Exploring a wide variety of ways to decompose the problem. Searching for general modularity (\emph{Q-Mod}) results in an intermediate level of population diversity. Using performance as the only objective results in the lowest level of structural diversity, since networks without any structural objective tend to become very densely connected, leaving all input neurons in the same module (Figure~\ref{fig:nns_retina_pa}).

Previous work on guiding evolution towards more modular networks on the retina problem has indicated that having an equally strong pressure on performance and structural objectives may lead evolution towards pathological, poorly performing structures~\citep{Clune2013}. We therefore also tested applying the modularity-objective \emph{probabilistically}, affecting selection only 25\% of the time, as proposed in~\citep{Clune2013}. While this does improve the performance of the general modularity-objective, both \emph{UserMod} and \emph{ModDiv} still reach the optimal solution faster (Supplementary Material Figure 1).


\subsubsection{The robot locomotion problem}

\begin{figure}[h]
	\centering
	\begin{subfigure}[b]{0.49\textwidth}
		\includegraphics[width=\textwidth]{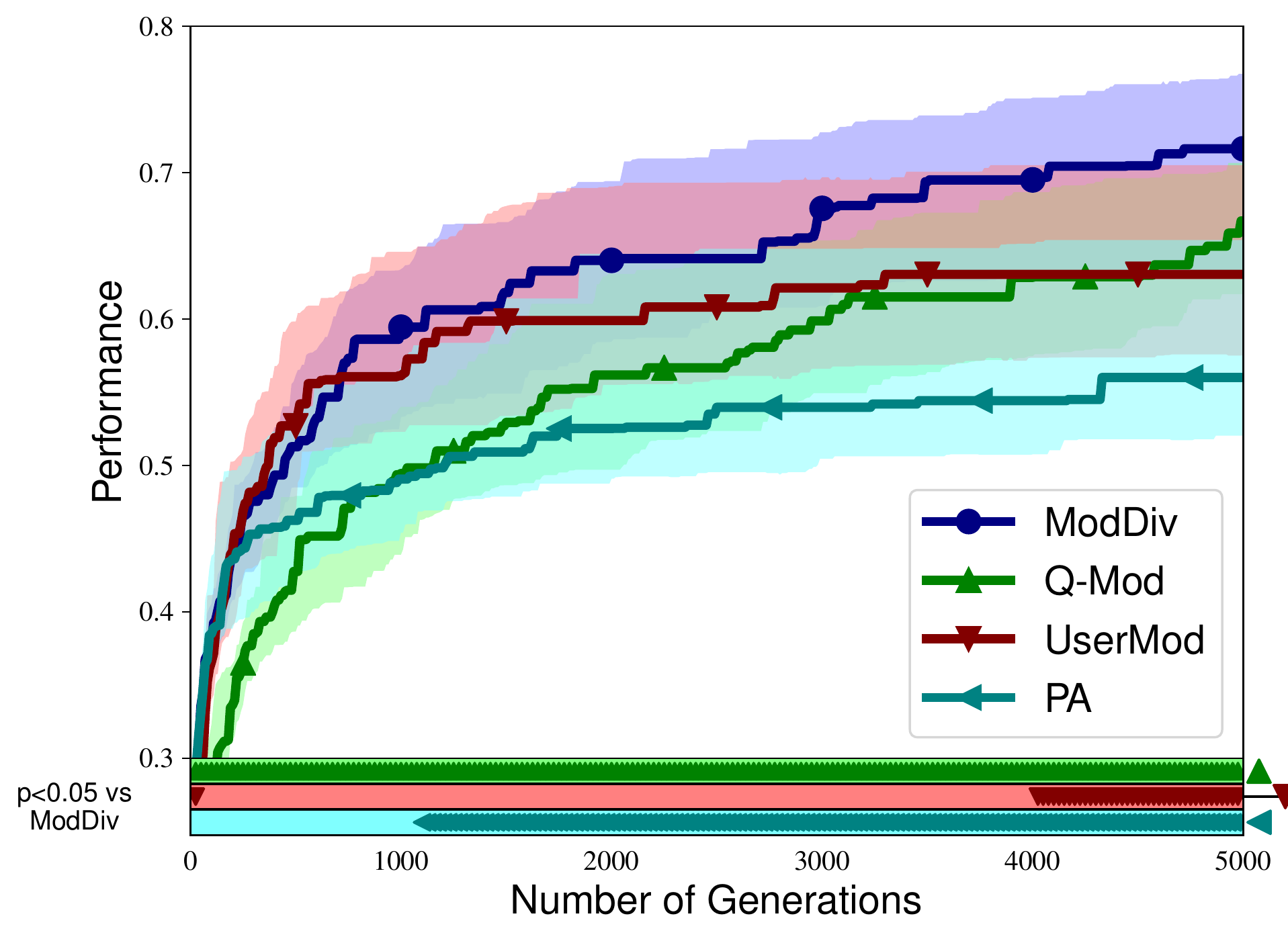}
		\caption{Performance of the best individual}
		\label{fig:spider_performance}
	\end{subfigure}
	\begin{subfigure}[b]{0.49\textwidth}
		\includegraphics[width=\textwidth]{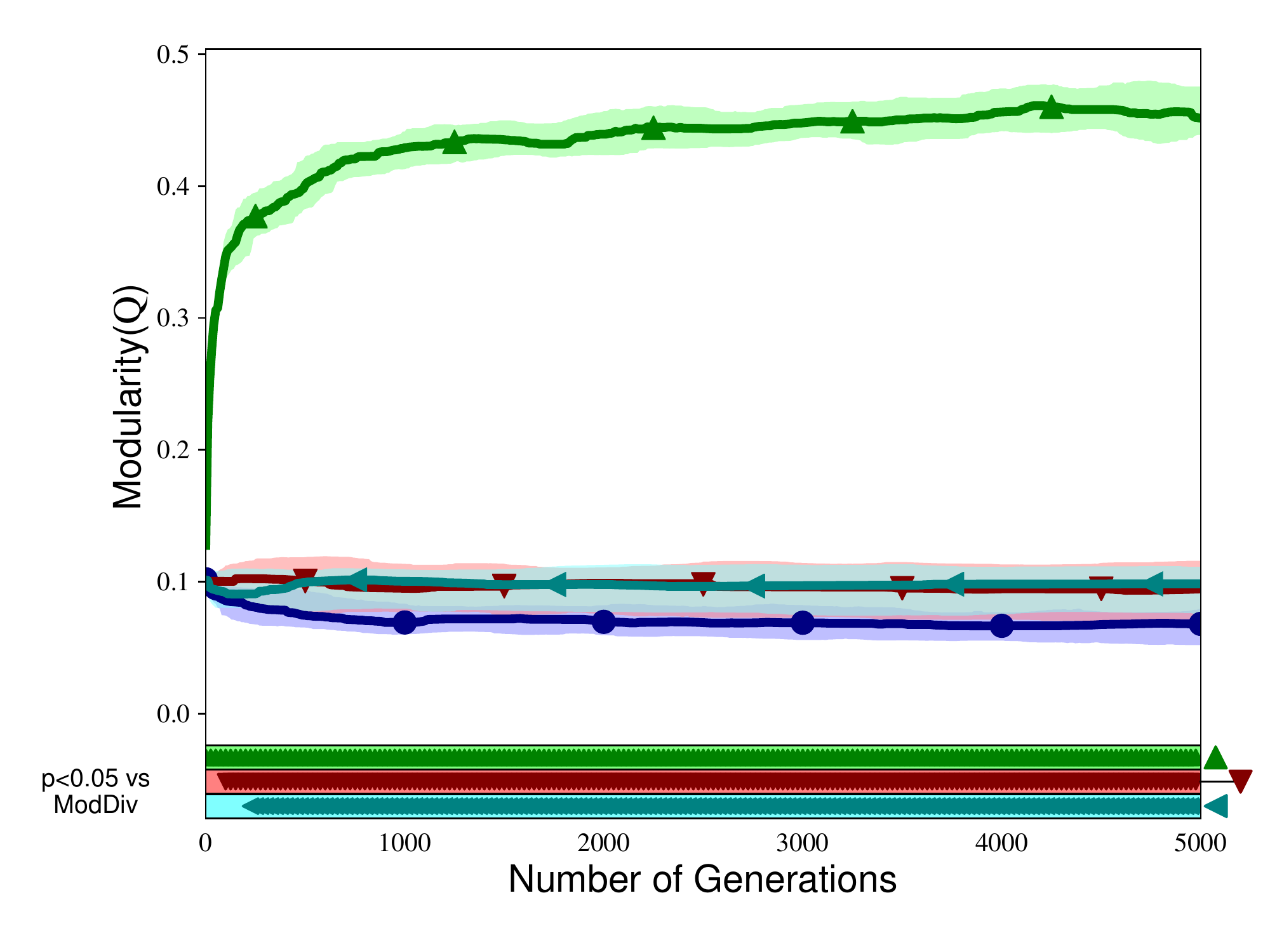}
		\caption{Median population modularity (Q-score)}
		\label{fig:spider_modularity}
	\end{subfigure}
	
	\begin{subfigure}[b]{0.49\textwidth}
		\includegraphics[width=\textwidth]{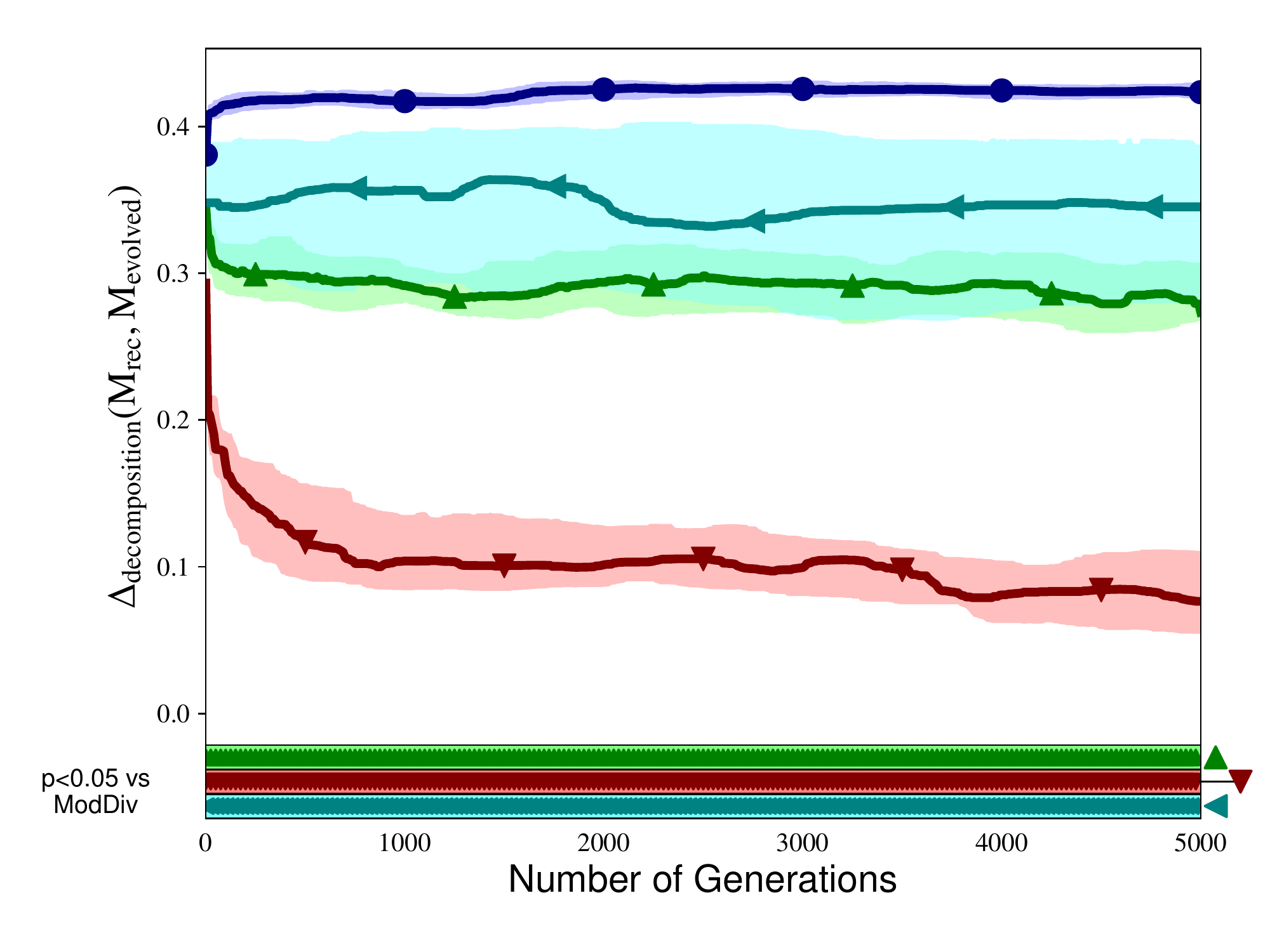}
		\caption{Median population distance from $M_{rec}$}
		\label{fig:spider_featuremod}
	\end{subfigure}
	\begin{subfigure}[b]{0.49\textwidth}
		\includegraphics[width=\textwidth]{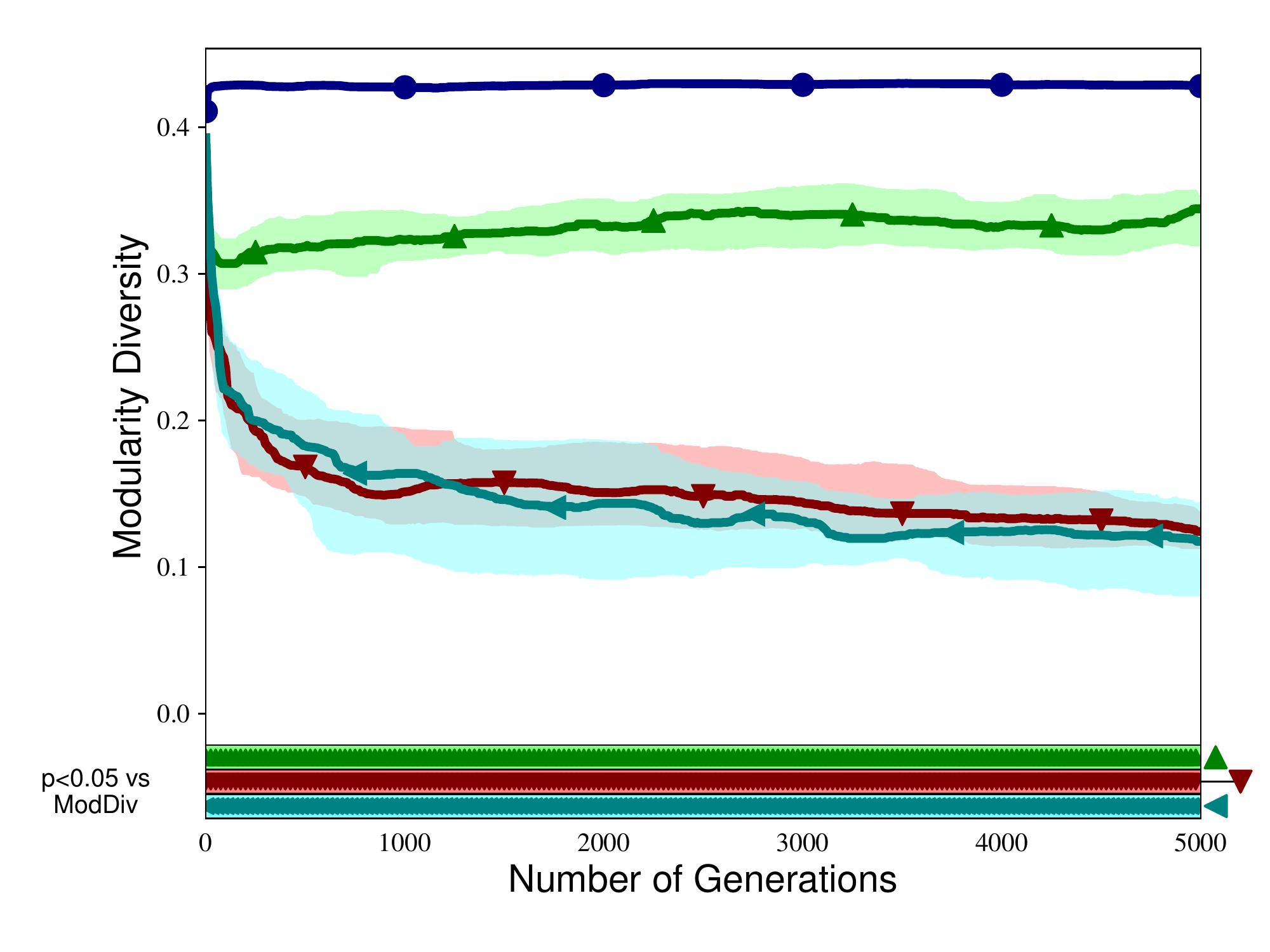}
		\caption{Median population diversity.}
		\label{fig:spider_diversity}
	\end{subfigure}
	\caption{The performance and development of different modular decompositions during evolution on the robotic locomotion problem. Modularity Diversity significantly outperforms the other treatments.}
	\label{fig:spider_result}
\end{figure}

The robot locomotion problem benefits the most from guidance by a structural 
diversity objective - which significantly outperforms all other treatments 
(Figure~\ref{fig:spider_performance}). The performance of the user-defined 
modularity pattern is weaker on this problem -- we believe the reason is that 
the problem has a less obviously modular structure. However, it cannot be 
ruled out that a different recommended decomposition could improve 
the performance of the \emph{UserMod} treatment. This highlights a limitation 
of the technique: It requires the user to correctly identify the right way to 
decompose the neural network. In agreement with 
the retina-problem, we see \emph{Q-Mod} producing the most modular structures 
(Figure~\ref{fig:spider_modularity}) and the same relative ordering of how 
diverse generated solutions are 
(Figure~\ref{fig:spider_diversity}).\footnote{Videos of the best and median 
resulting robot gaits across 50 replications of each treatment can be seen at 
\url{https://youtu.be/ZbP1JgQffLI} and \url{https://youtu.be/cPS-7g65YwY}, 
respectively.}

\subsection{Neural network structures}

In this section, we show and analyze the structure of final evolved networks. The presented ANNs are all ``winners'' of their respective evolutionary run, meaning they performed the best on the primary objective and, in case of ties, outperformed others on the secondary objective. We focus on median results from the 50 repetitions of each treatment, since they reveal the most interesting differences: All treatments \emph{occasionally} reach very good performance -- the main advantage of structural guidance is that very good performance is reached much more frequently.

\subsubsection{The retina problem}
\begin{figure}
\centering

\begin{subfigure}[b]{\textwidth}
	
	\includegraphics[width=0.24\textwidth]{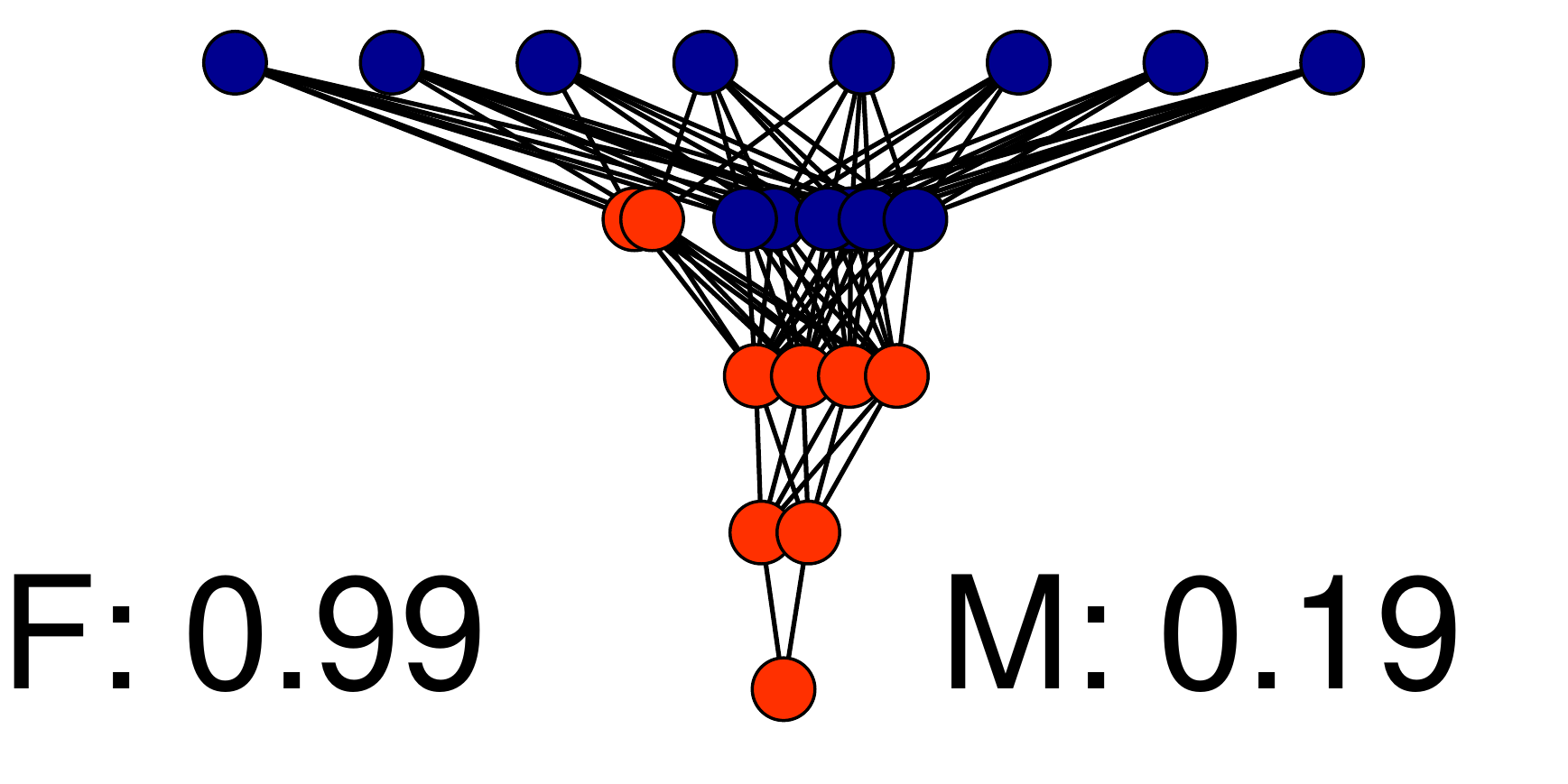}	
	\includegraphics[width=0.24\textwidth]{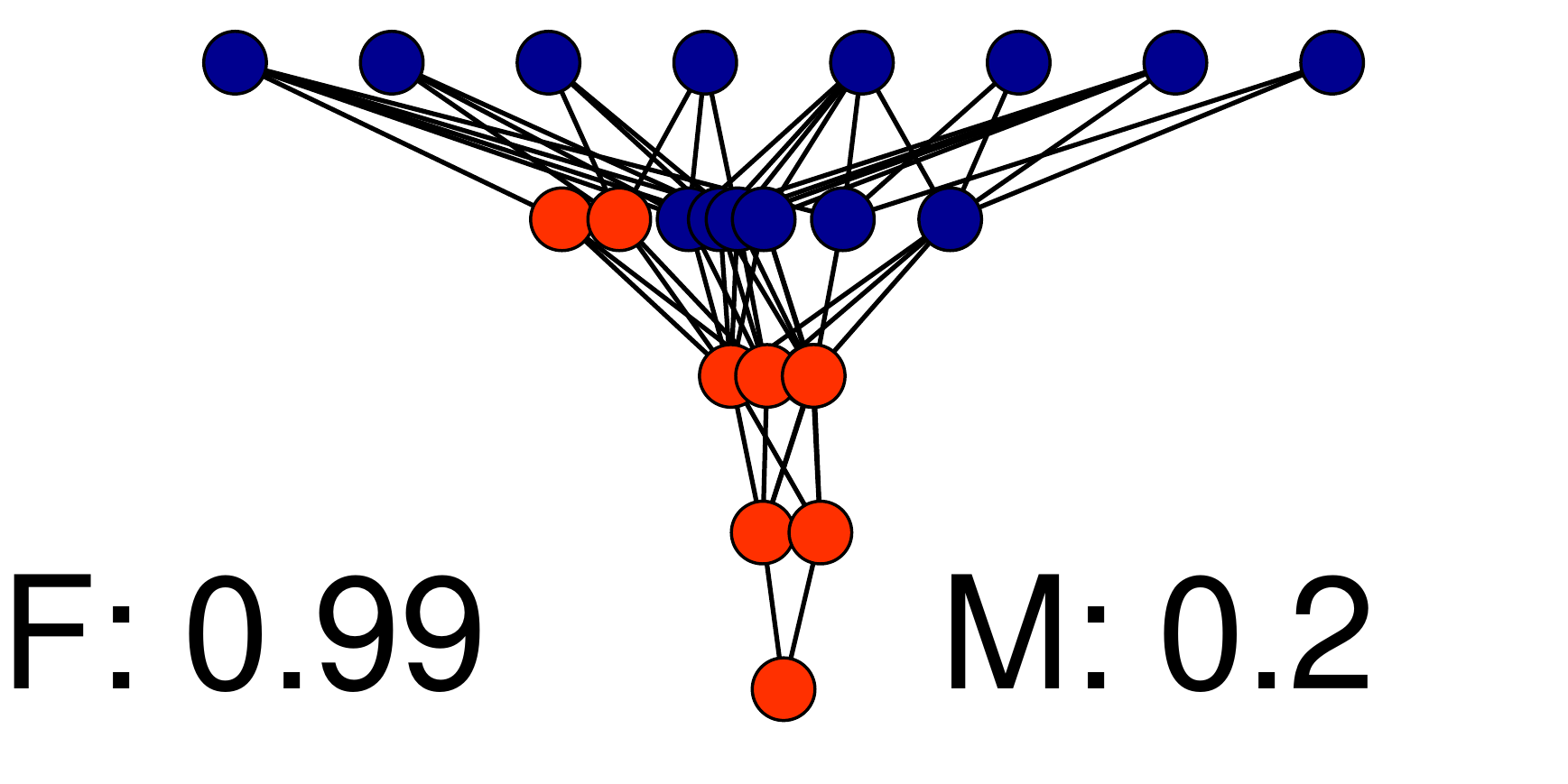}	
	\includegraphics[width=0.24\textwidth]{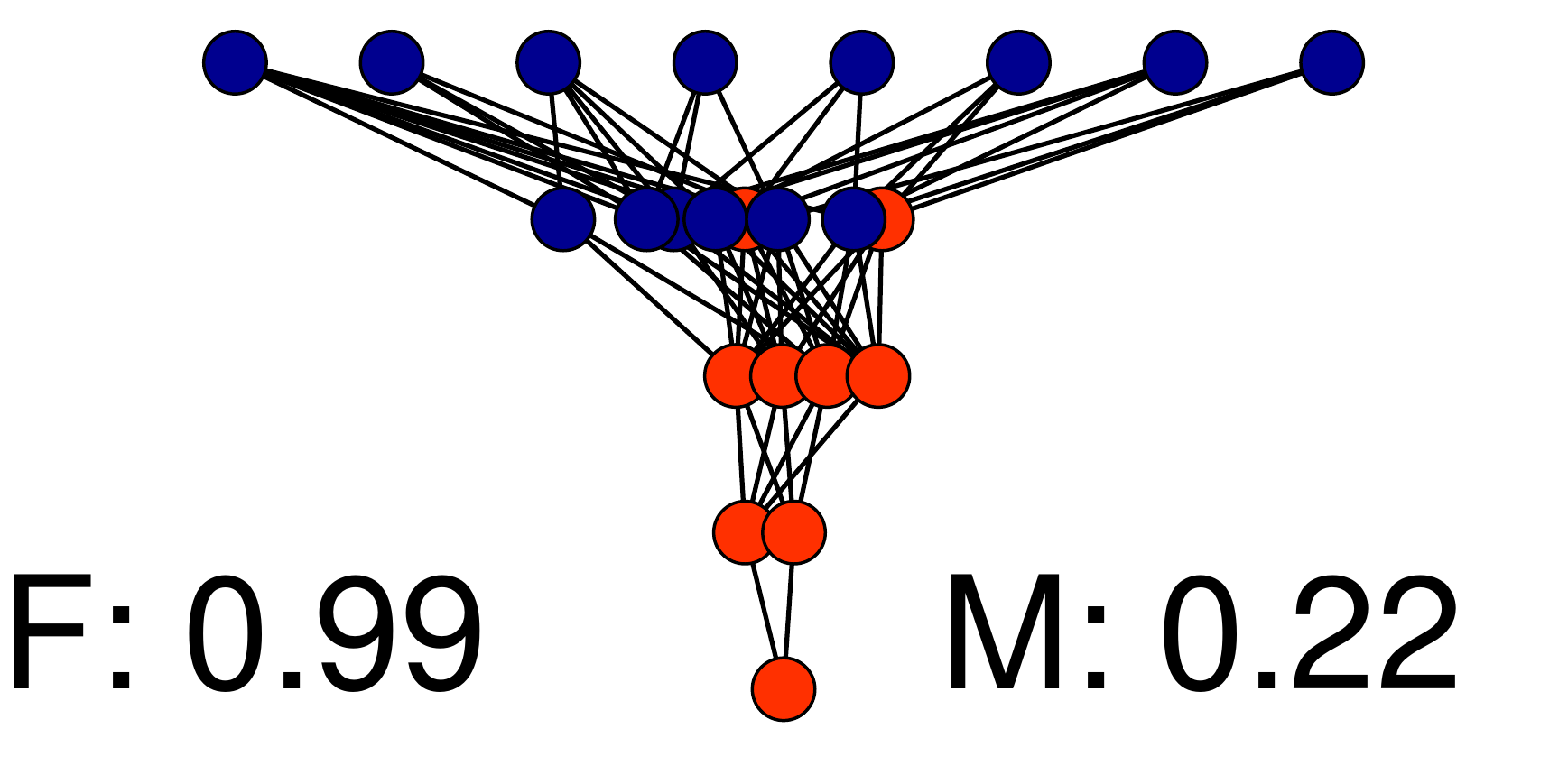}	
	\includegraphics[width=0.24\textwidth]{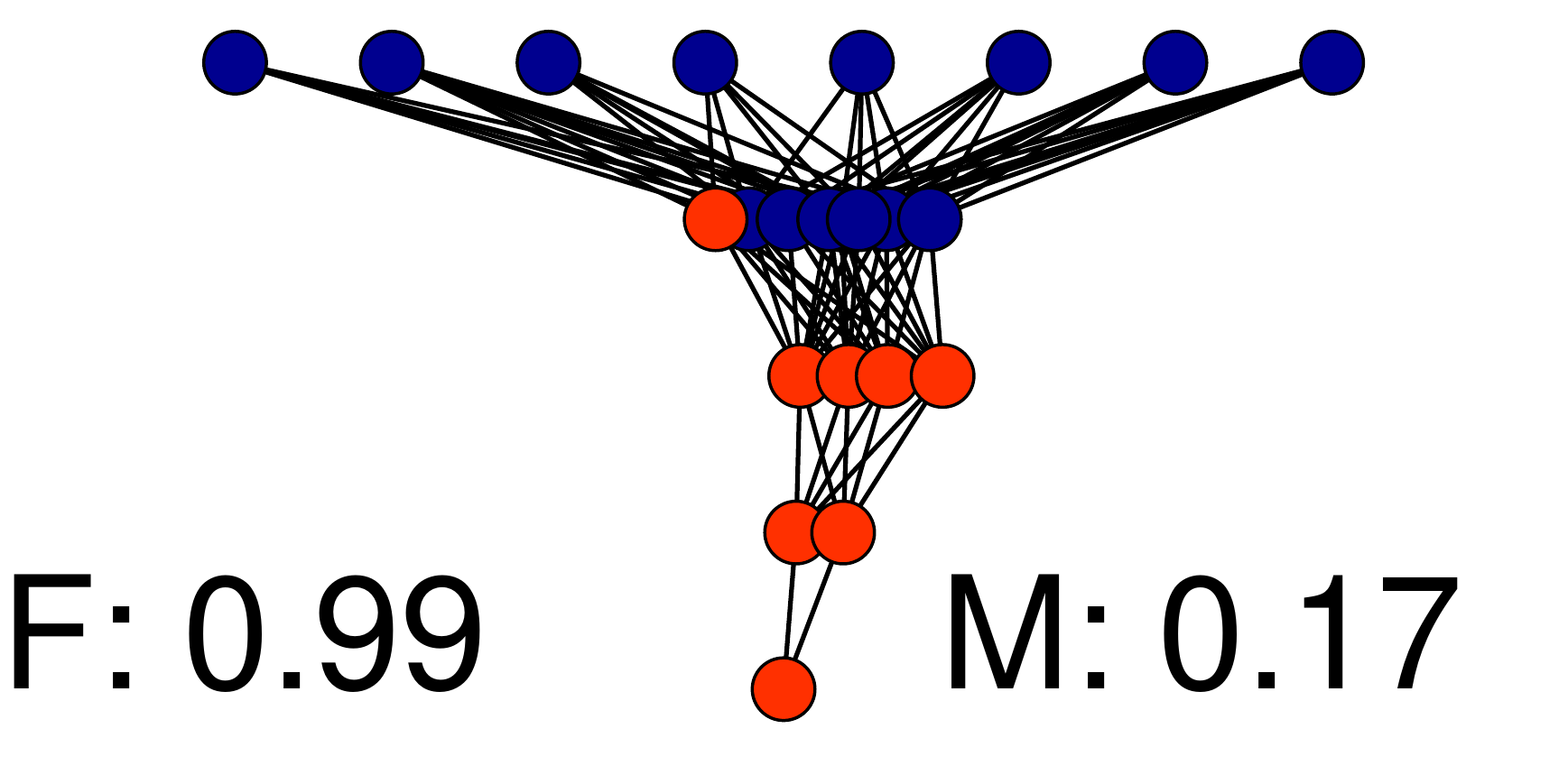}
	\caption{Performance Alone (no structural objective)}
	\label{fig:nns_retina_pa}
\end{subfigure}

\begin{subfigure}[b]{\textwidth}
	
	\includegraphics[width=0.24\textwidth]{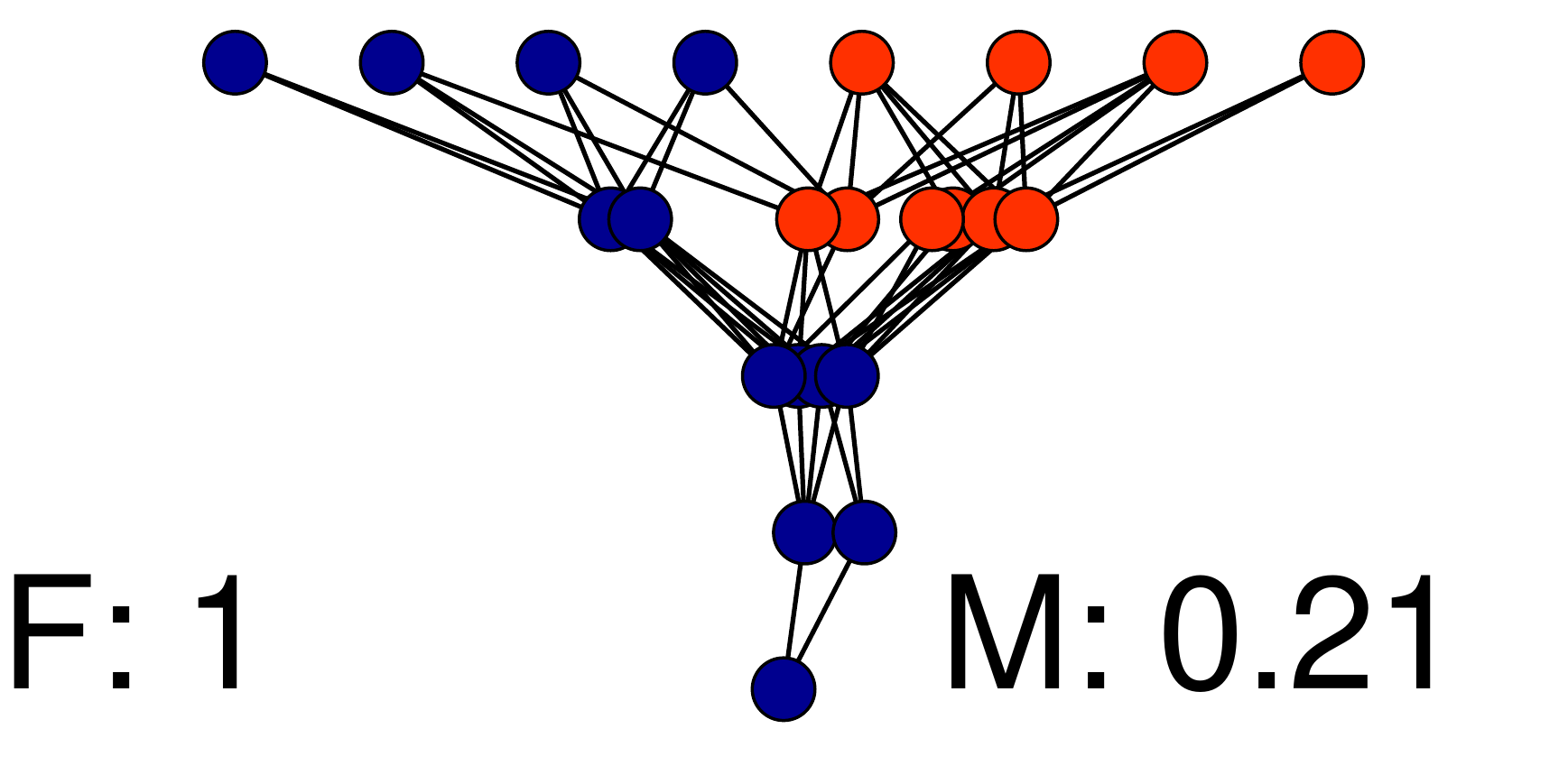}	
	\includegraphics[width=0.24\textwidth]{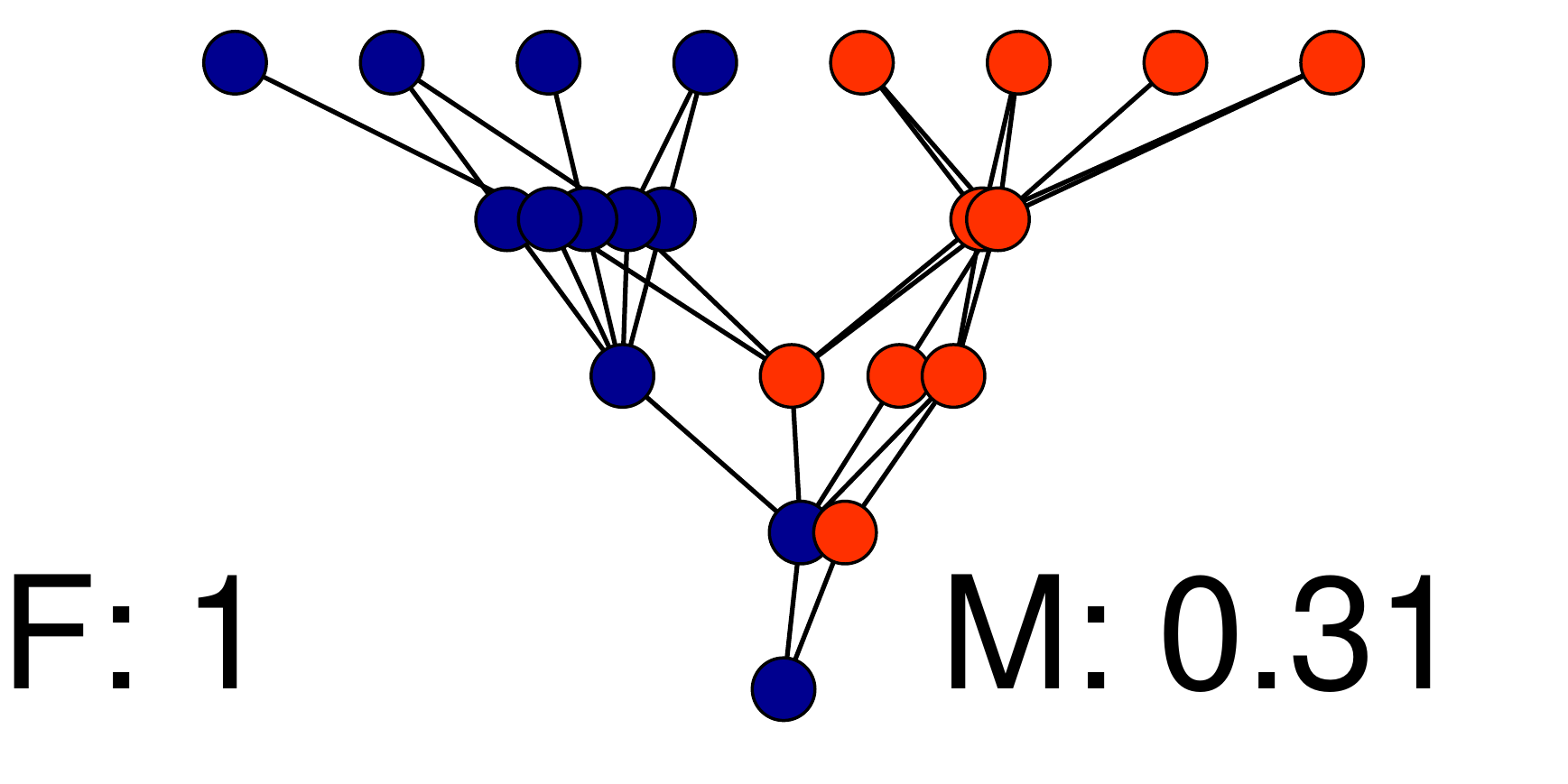}	
	\includegraphics[width=0.24\textwidth]{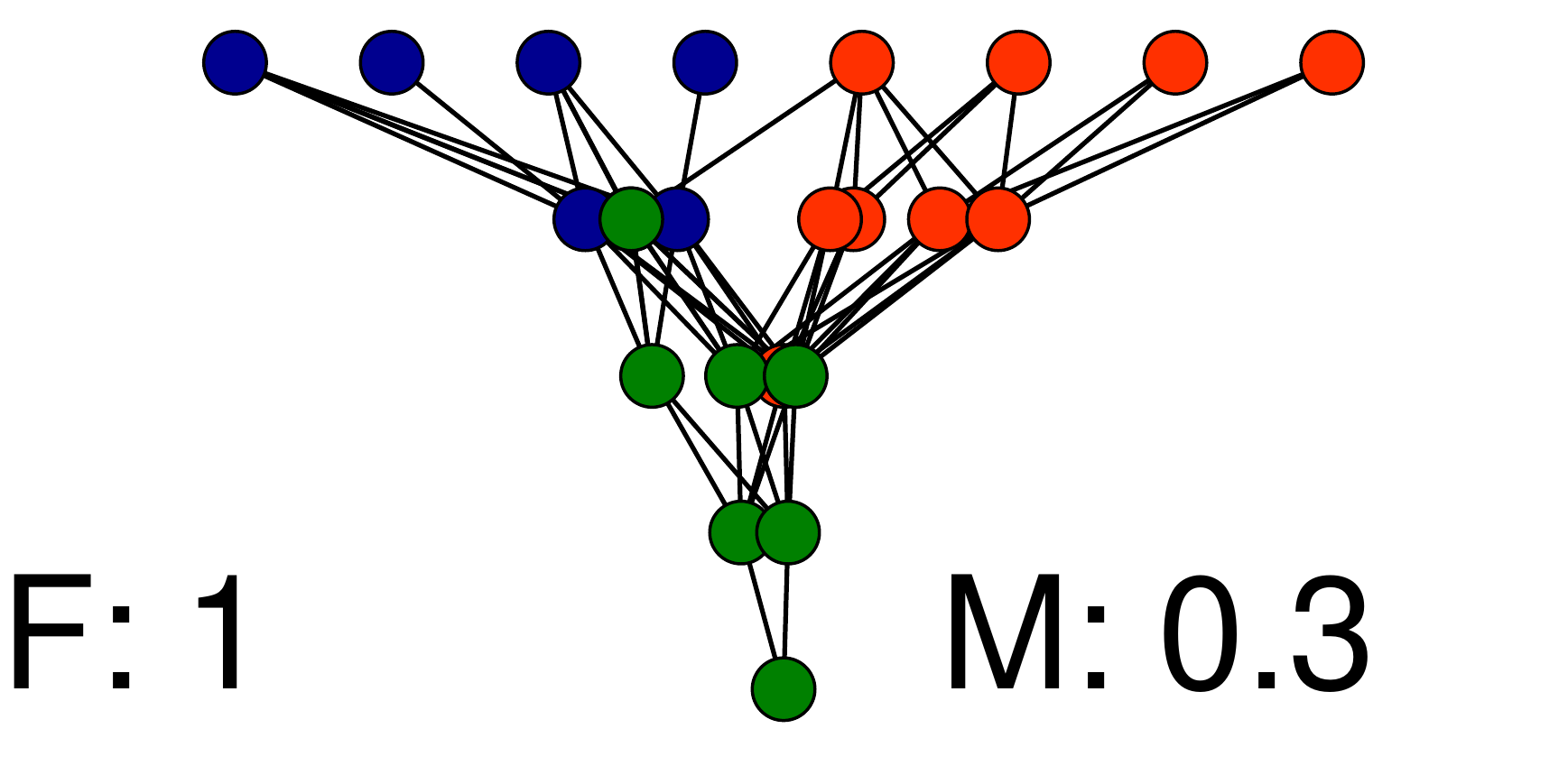}	
	\includegraphics[width=0.24\textwidth]{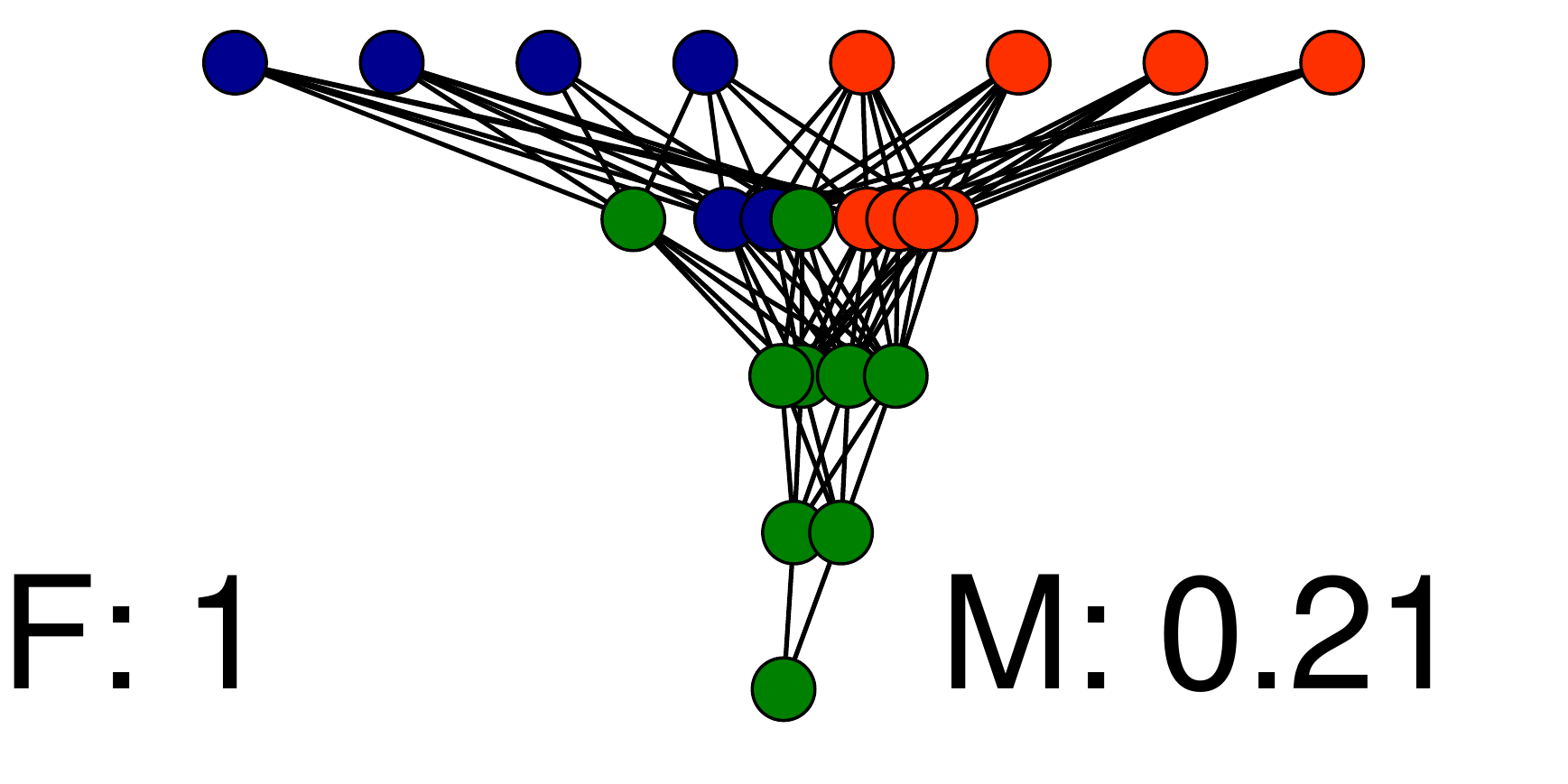}
	\caption{User-defined Modularity}
	\label{fig:nns_retina_usermod}
\end{subfigure}

\begin{subfigure}[b]{\textwidth}
	
	\includegraphics[width=0.24\textwidth]{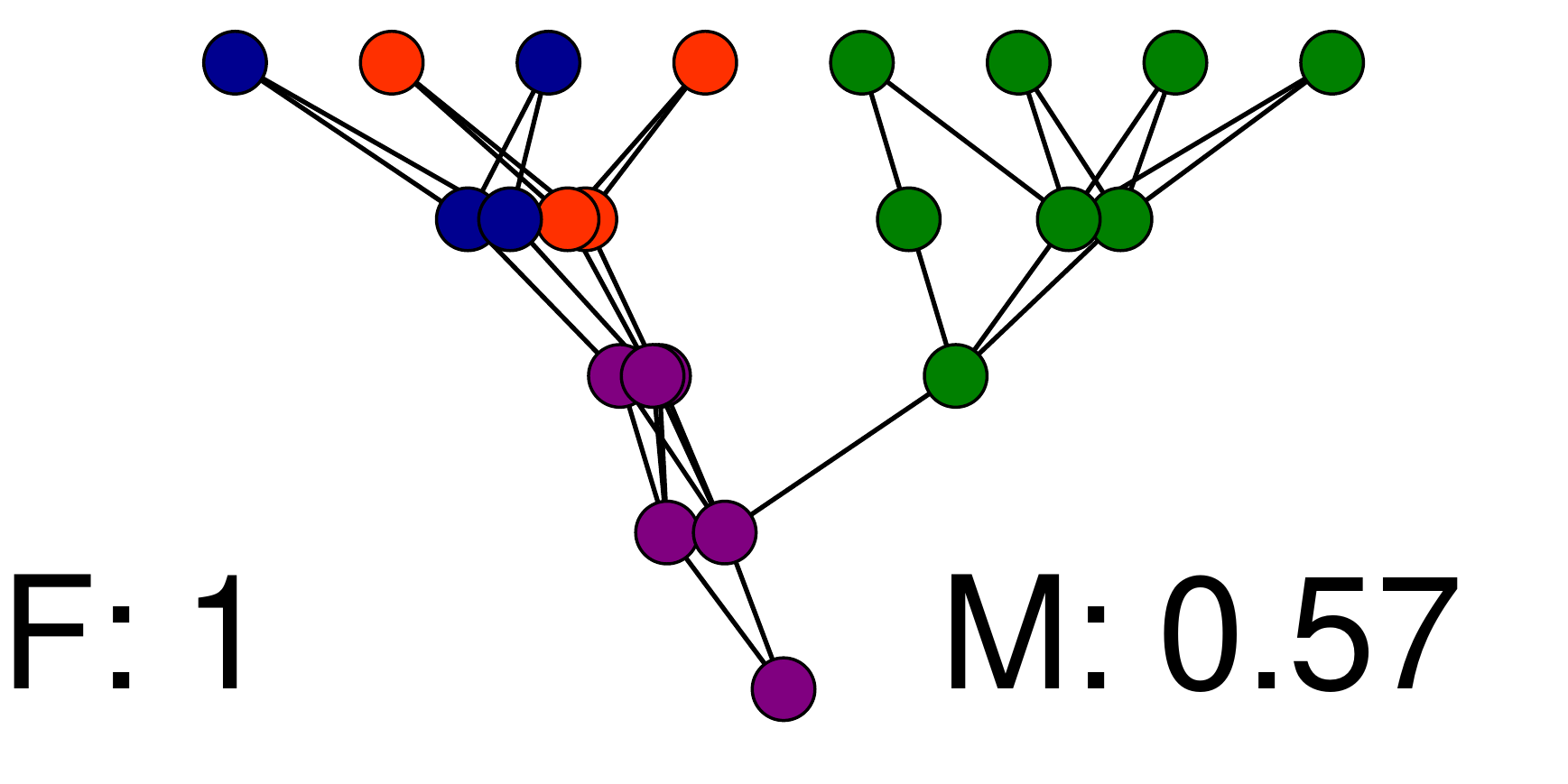}	
	\includegraphics[width=0.24\textwidth]{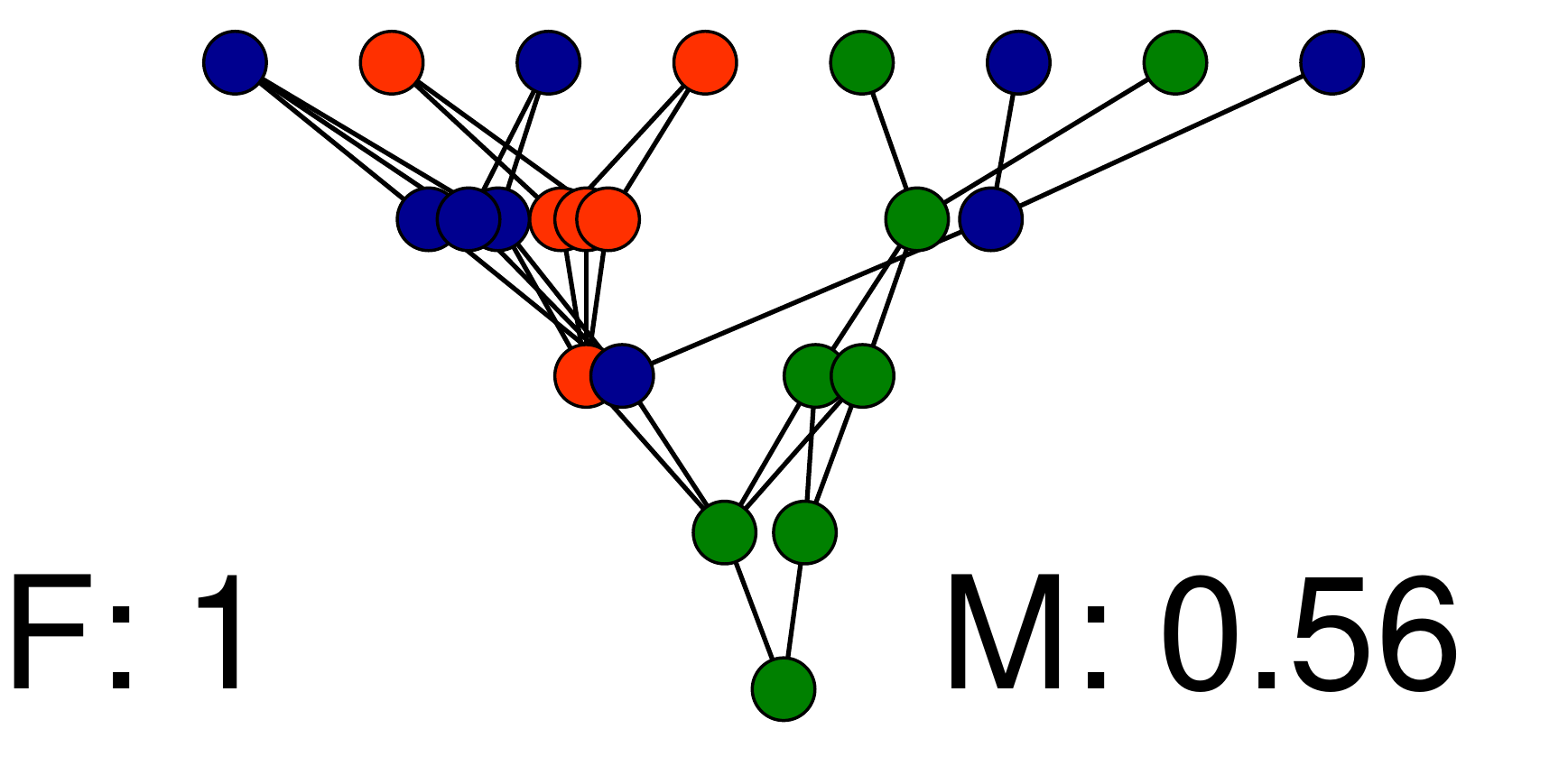}	
	\includegraphics[width=0.24\textwidth]{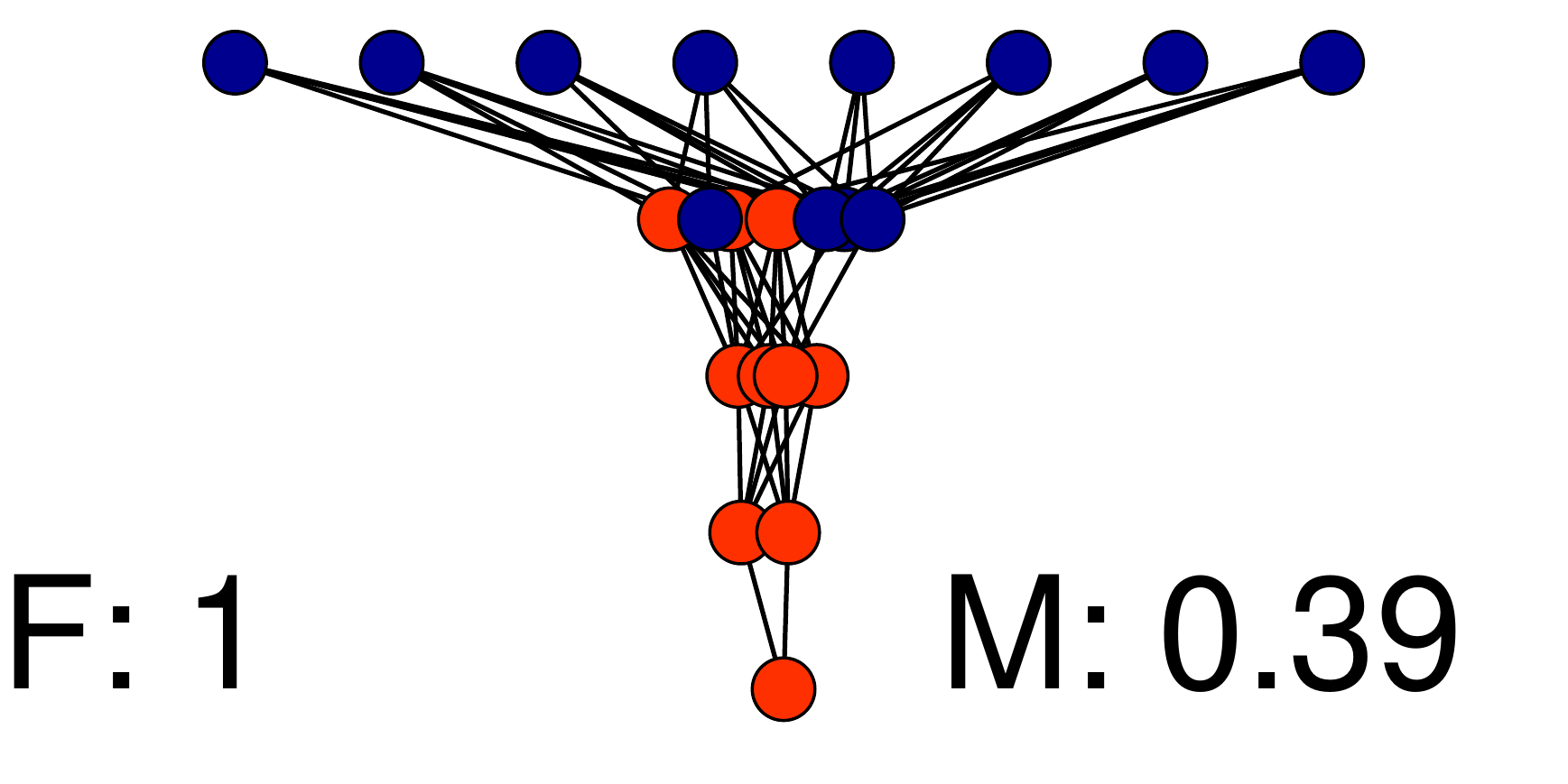}	
	\includegraphics[width=0.24\textwidth]{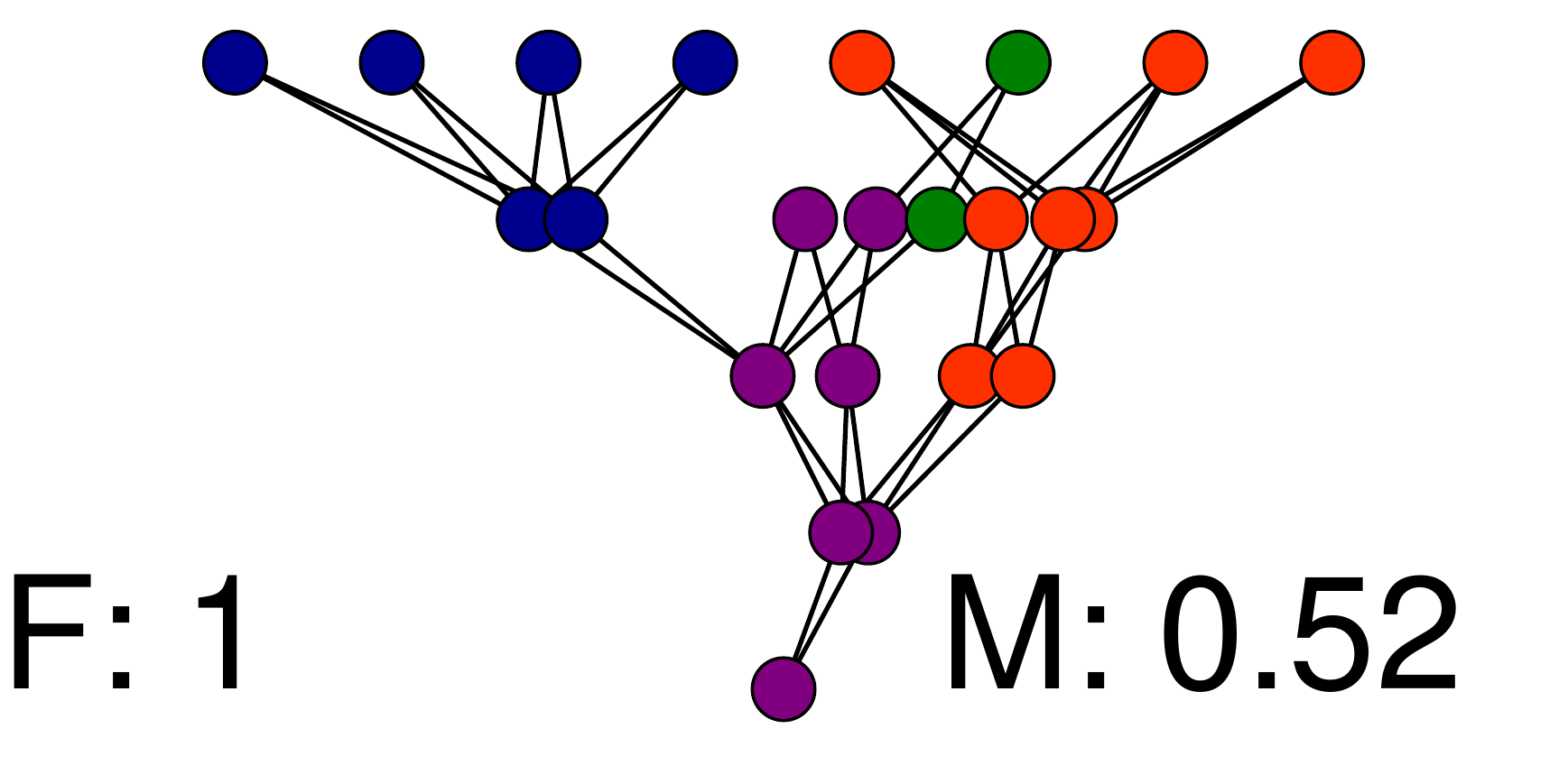}
	\caption{General Modularity}
	\label{fig:nns_retina_qmod}
\end{subfigure}

\begin{subfigure}[b]{\textwidth}
	\includegraphics[width=0.24\textwidth]{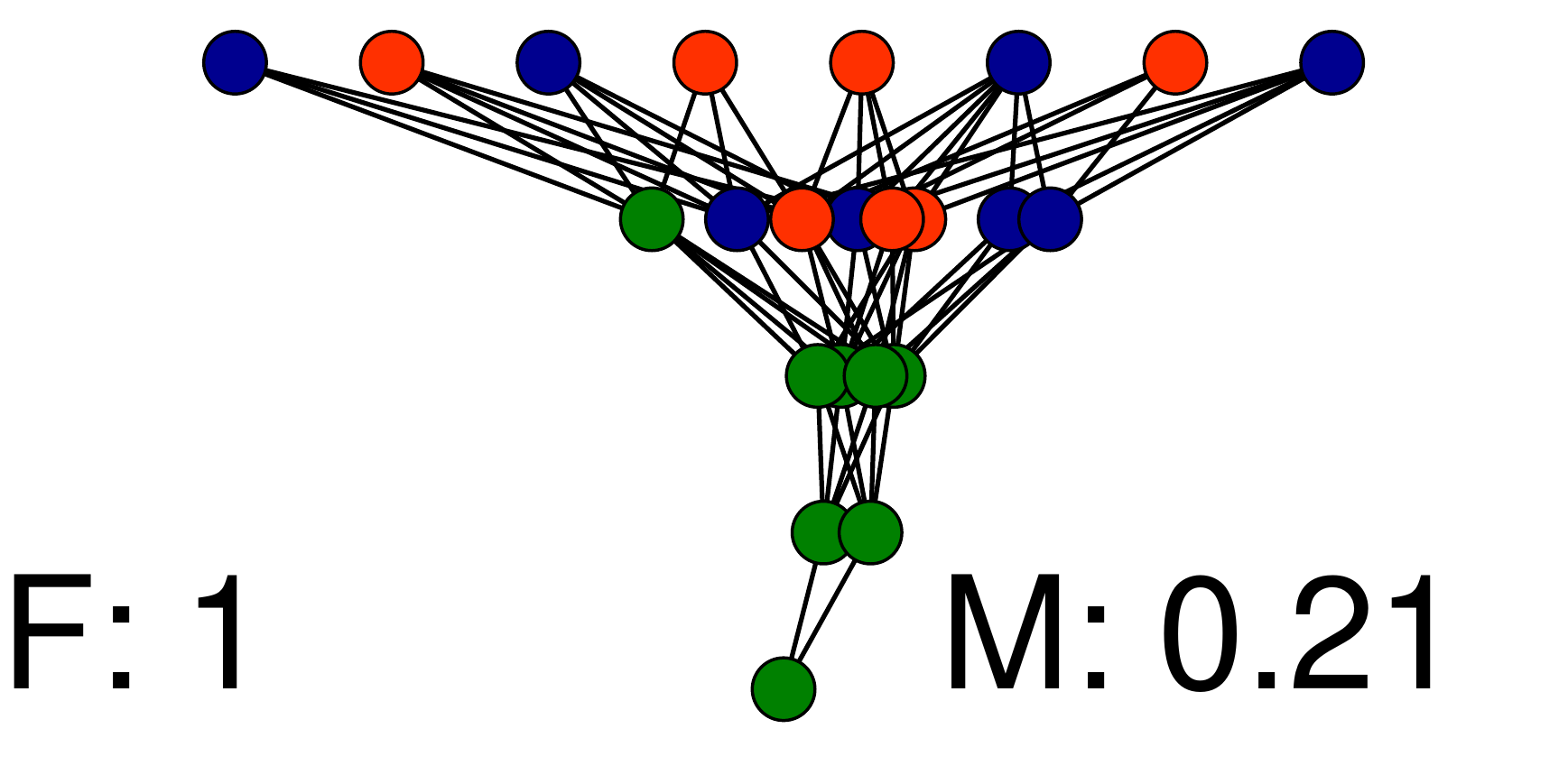}	
	\includegraphics[width=0.24\textwidth]{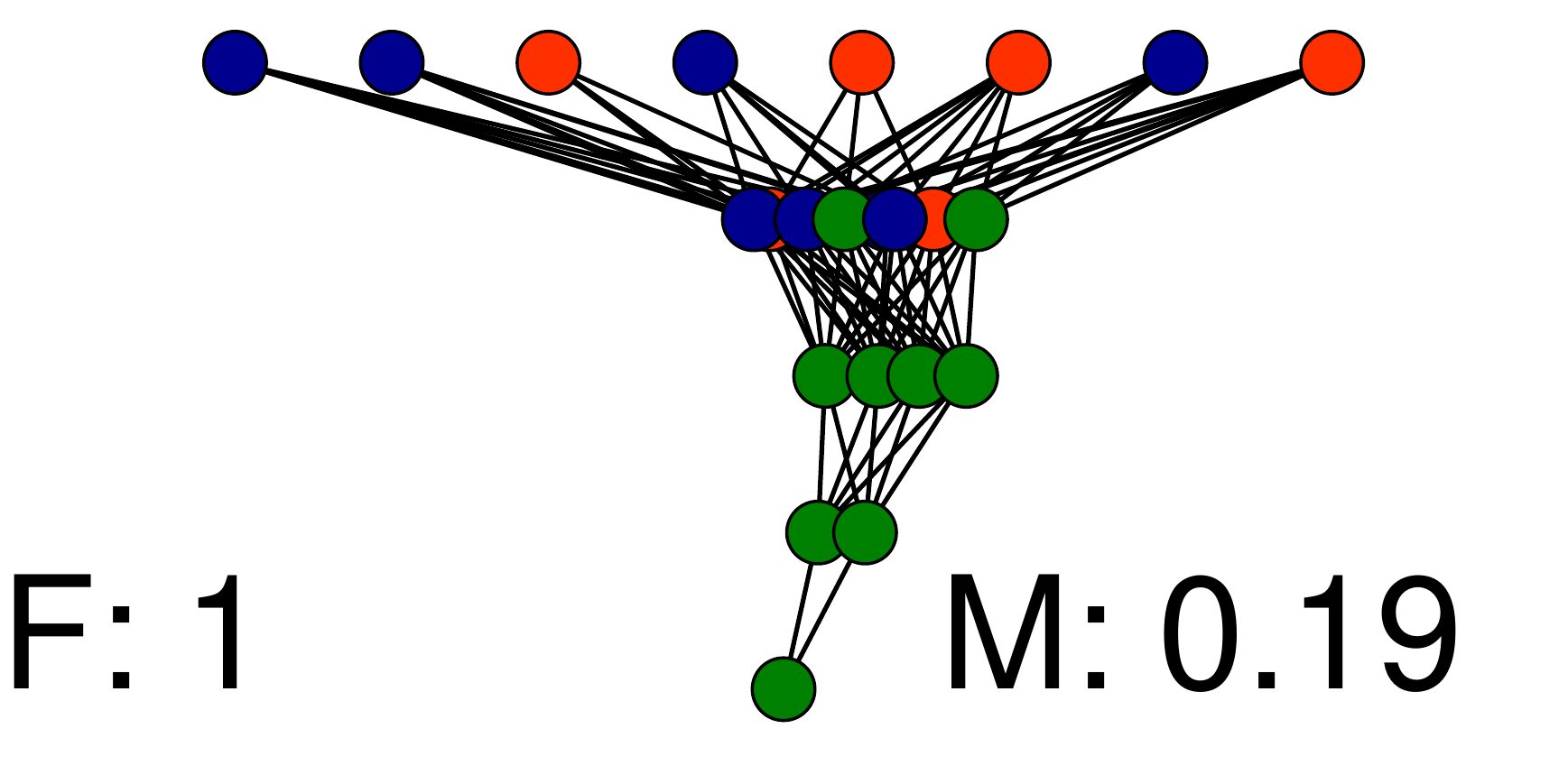}	
	\includegraphics[width=0.24\textwidth]{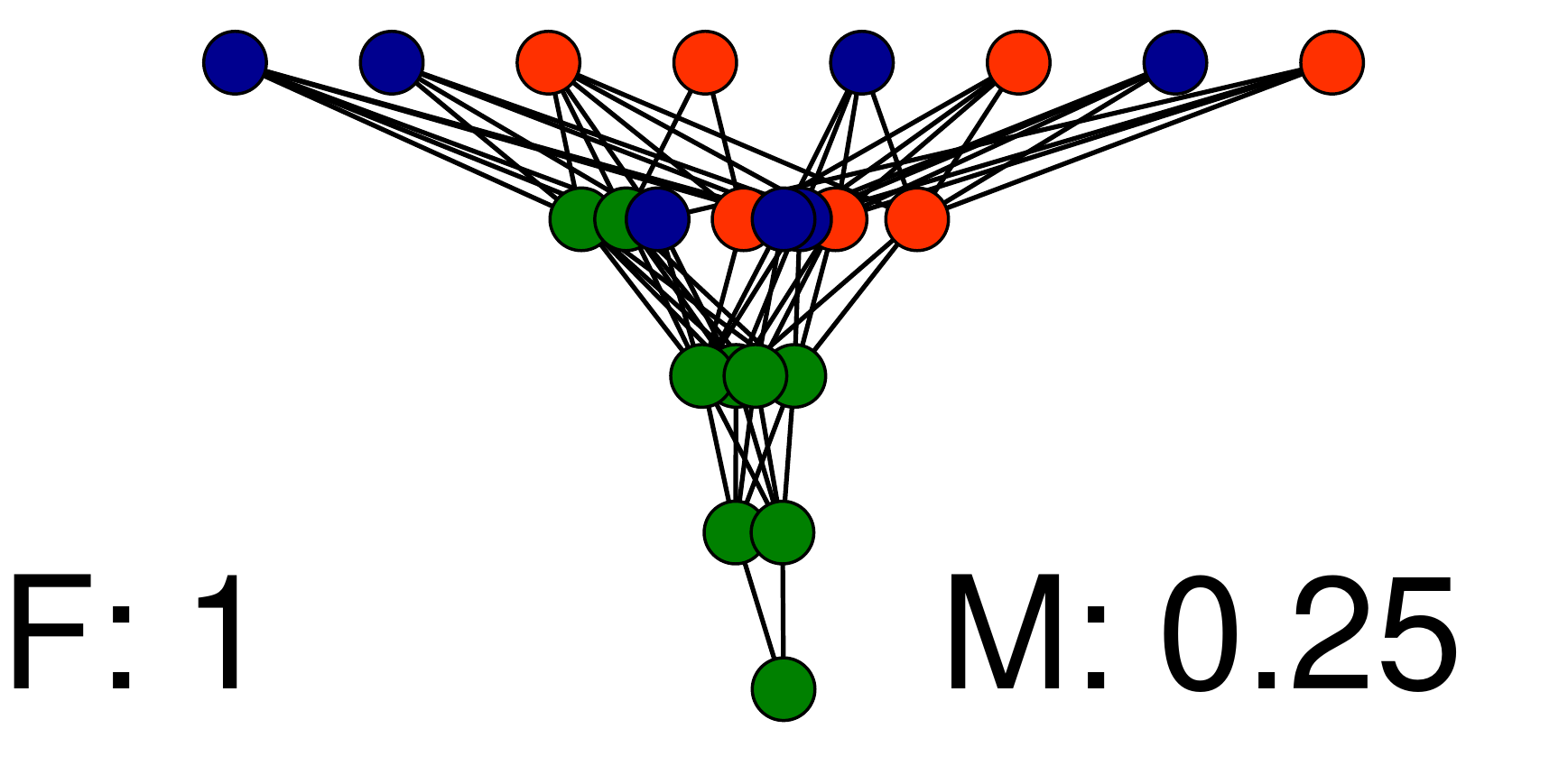}	
	\includegraphics[width=0.24\textwidth]{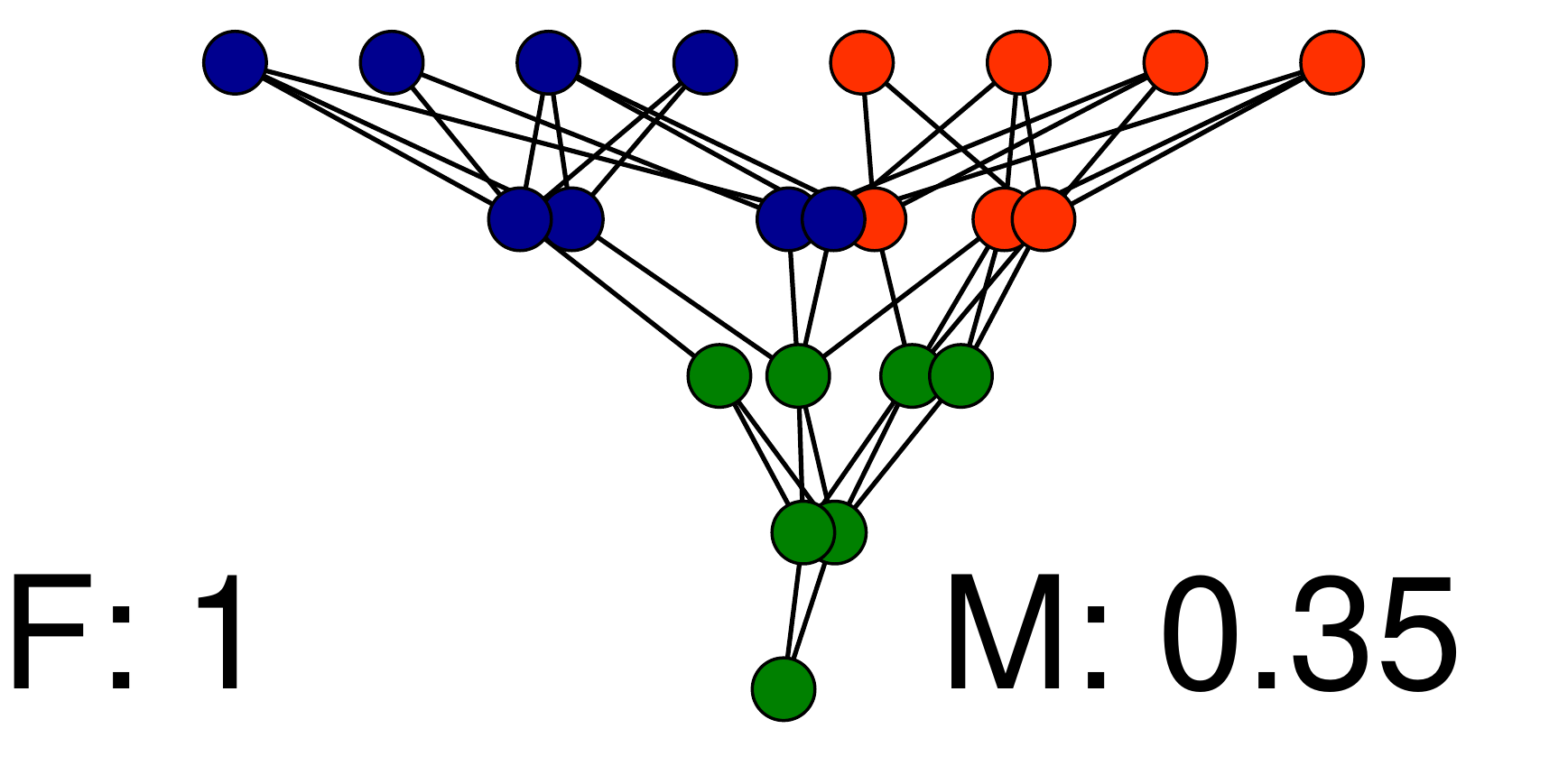}
	\caption{Modularity Diversity}
	\label{fig:nns_retina_div}
\end{subfigure}

	\caption{The structure of typical (median) neural networks evolved for the retina task. All 50 winner networks are shown in the Supplementary Material. F: Fitness on the retina-task. M: Modularity Q-score.}
	\label{fig:nns_retina}
\end{figure}

Figure~\ref{fig:nns_retina} shows final evolved networks for the retina-problem. The Performance Alone treatment results in entangled networks without any modular structure on input neurons (Figure~\ref{fig:nns_retina_pa}). Both general modularity (\emph{Q-Mod}) and user-defined modularity as a guiding objective result in networks frequently matching the recommended problem decomposition (Figures~\ref{fig:nns_retina_usermod} and~\ref{fig:nns_retina_qmod}) -- but as one might expect, \emph{UserMod} tends to do so more frequently (92\% of \emph{UserMod} networks match the recommended decomposition, vs 22\% of \emph{Q-Mod} networks). Guiding evolution towards a diverse collection of modularity patterns has the effect of producing networks with unexpected, yet well-performing problem decompositions (Figure~\ref{fig:nns_retina_div}).


\subsubsection{The robot locomotion problem}

The neural networks evolved for the robot locomotion problem reveal interesting properties about the applied structural objectives. Guiding evolution with the user-defined modularity pattern from Figure~\ref{fig:spider_decomposition} results in 42\% of the winner-networks perfectly matching the recommended structure (Supplementary Material Figure 9-10). On this problem, guiding evolution with general modularity as the objective (\emph{Q-Mod}) \emph{never} results in reaching the recommended pattern (Supplementary Material Figure 11-12). Still, the \emph{performance scores} of \emph{Q-Mod} and \emph{UserMod} are comparable. This indicates that there are more alternative decomposition patterns to exploit for this problem -- there is a less clear relationship between modularity patterns and performance. Again we see the networks guided by Modularity Diversity outperform the others by reaching unexpected, well-performing decompositions (Supplementary Material Figure 13-14). 

\subsection{Structural versus behavioral diversity}

\begin{figure}
\centering
\begin{subfigure}[b]{0.48\textwidth}
	\includegraphics[width=\textwidth]{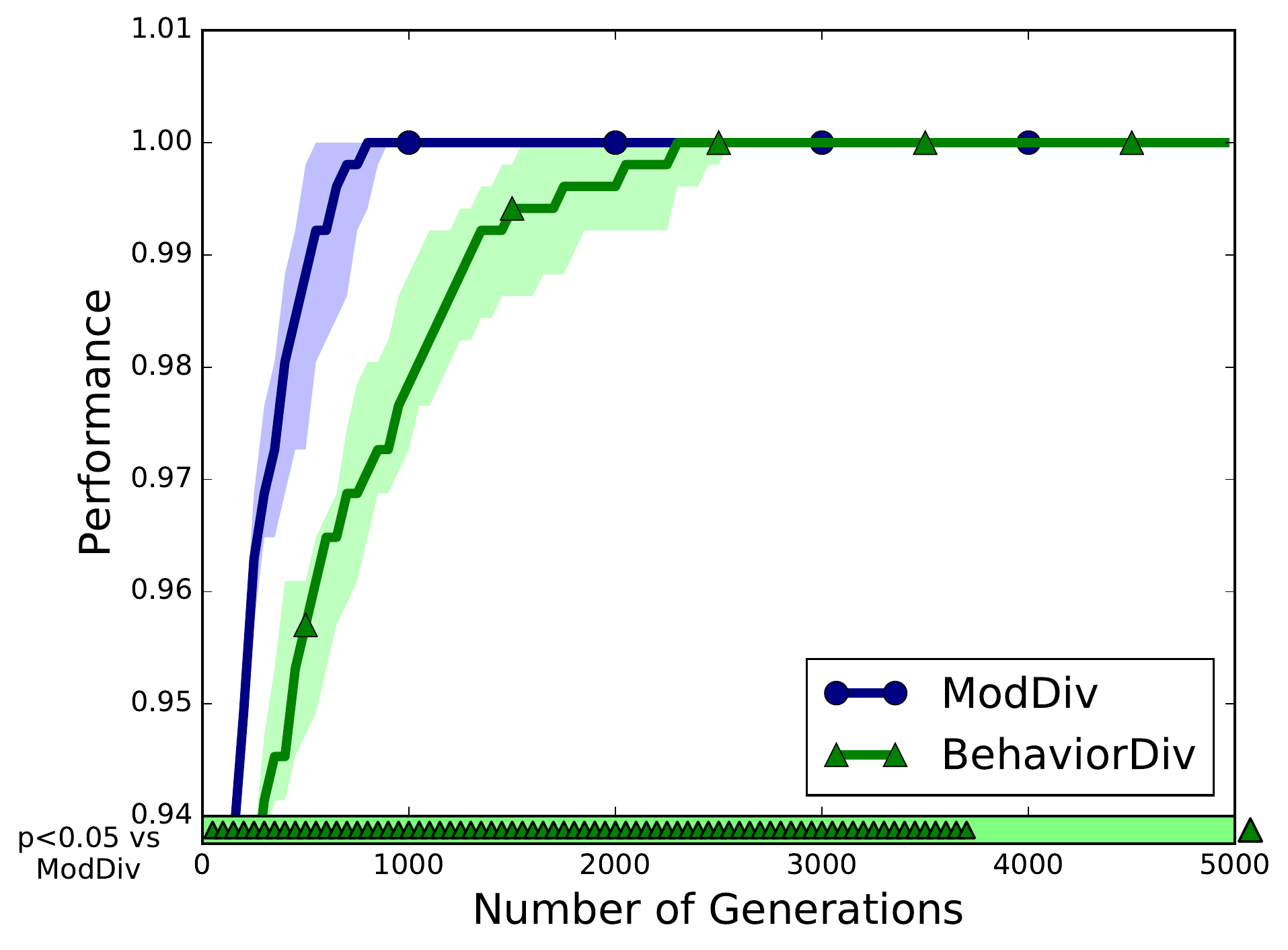}
	\caption{Performance on the retina problem}
	\label{fig:diversity_retina}
\end{subfigure}
\begin{subfigure}[b]{0.48\textwidth}
	\includegraphics[width=\textwidth]{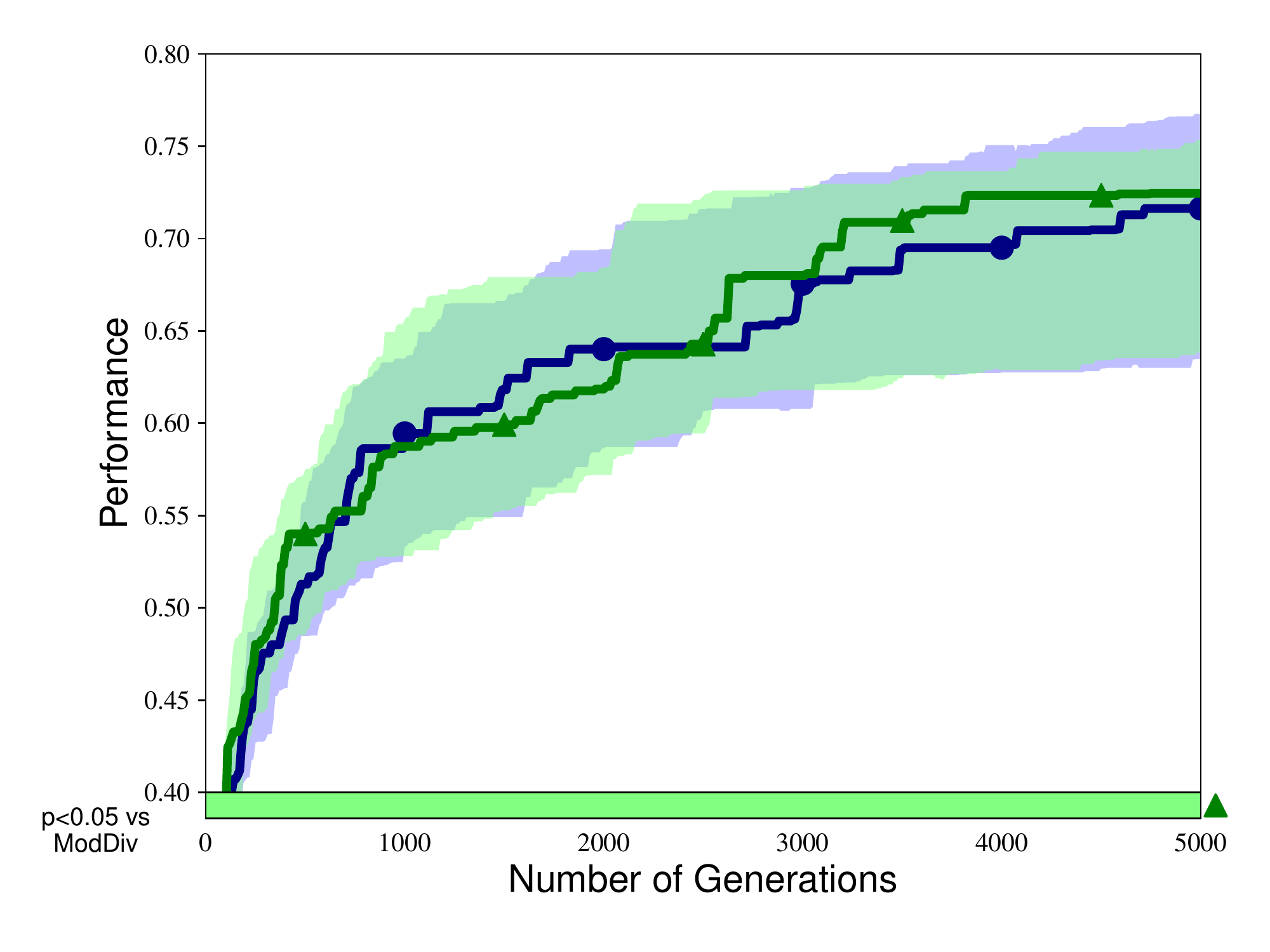}
	\caption{Performance on the locomotion problem}
	\label{fig:diversity_arm}
\end{subfigure}
	\caption{Comparing the result of encouraging structural versus behavioral diversity.}
	\label{fig:diversity_performance}
\end{figure}

To test how our structural diversity technique compares to the powerful technique of encouraging behavioral diversity, we evolved neural networks with the behavioral diversity measurement outlined in Section~\ref{sec:diversity_measurement} as a guiding objective, and compared the results to networks evolved with the Modularity Diversity objective. Figure~\ref{fig:diversity_performance} shows the resulting performance on the retina and robot locomotion task. For the retina, structural diversity leads to the best-performing solutions significantly faster than behavioral diversity. For the locomotion-problem, performance of the two is similar.

We hypothesize that the reason structural diversity does not outperform behavioral diversity on the robotic locomotion problem is that for this problem, the \emph{structure} of evolving networks is not very indicative of their potential performance -- this is supported by the finding that a general pressure towards modular networks \emph{never} leads to the user-recommended structure for this problem (Supplementary Material Figure 11-12). For the retina problem, structure and performance are more closely related, as indicated by the fact that the user-defined structure is reached almost immediately for the UserMod-treatment on this problem (Figure~\ref{fig:usermod_retina}). This makes a population rich in structural diversity the best guiding objective for the retina problem.

We consider it an important topic of future research to work towards a better understanding of which kind of problems can gain the most from guidance from high-level structural objectives.

\subsection{A non-modular problem}

\begin{figure}[h]
\centering
\begin{subfigure}[b]{0.49\textwidth}
	\includegraphics[width=\textwidth]{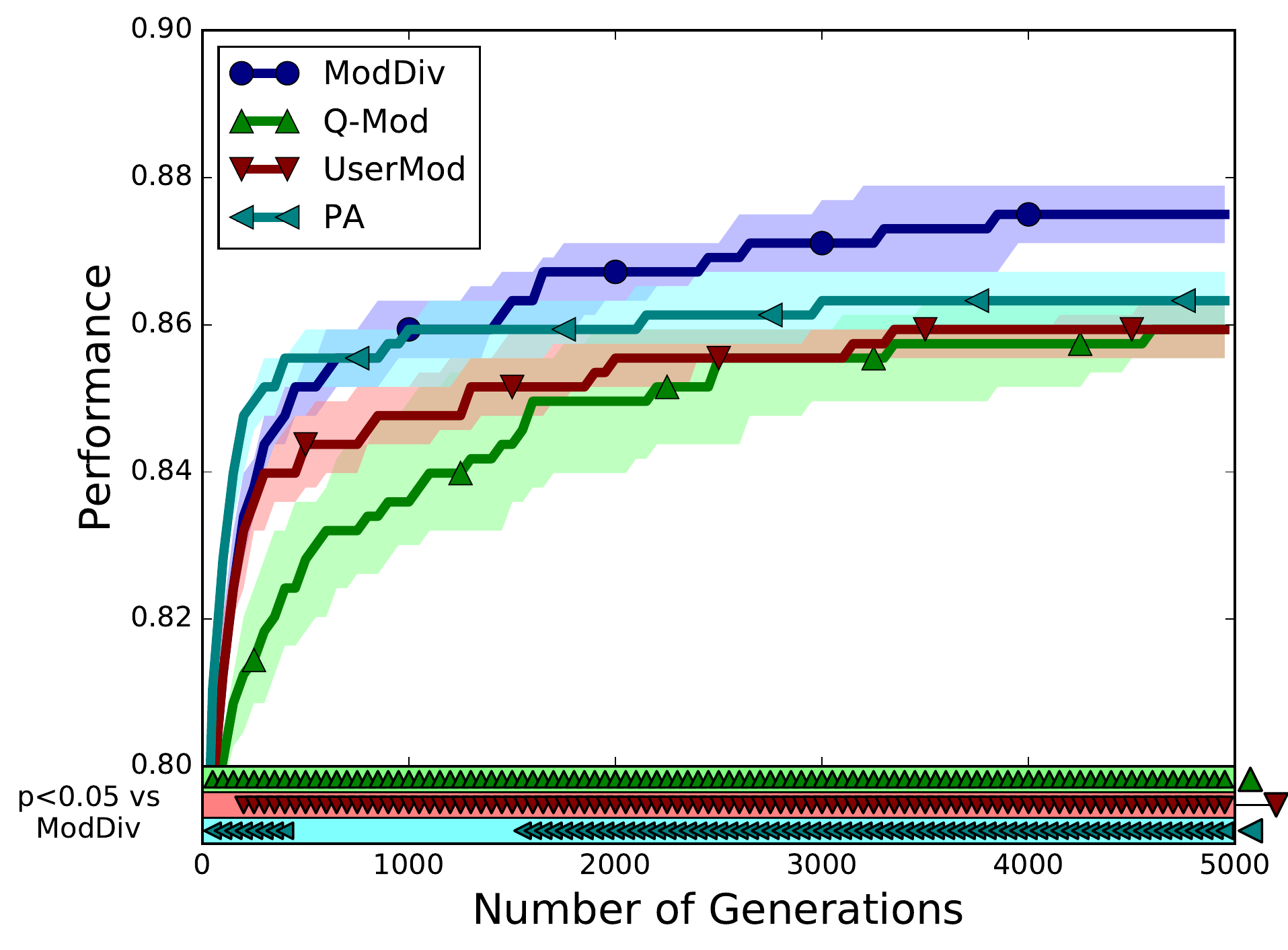}
	\caption{Performance of the best individual}
	\label{fig:nonmod_retina_performance}
\end{subfigure}
\begin{subfigure}[b]{0.49\textwidth}
	\includegraphics[width=\textwidth]{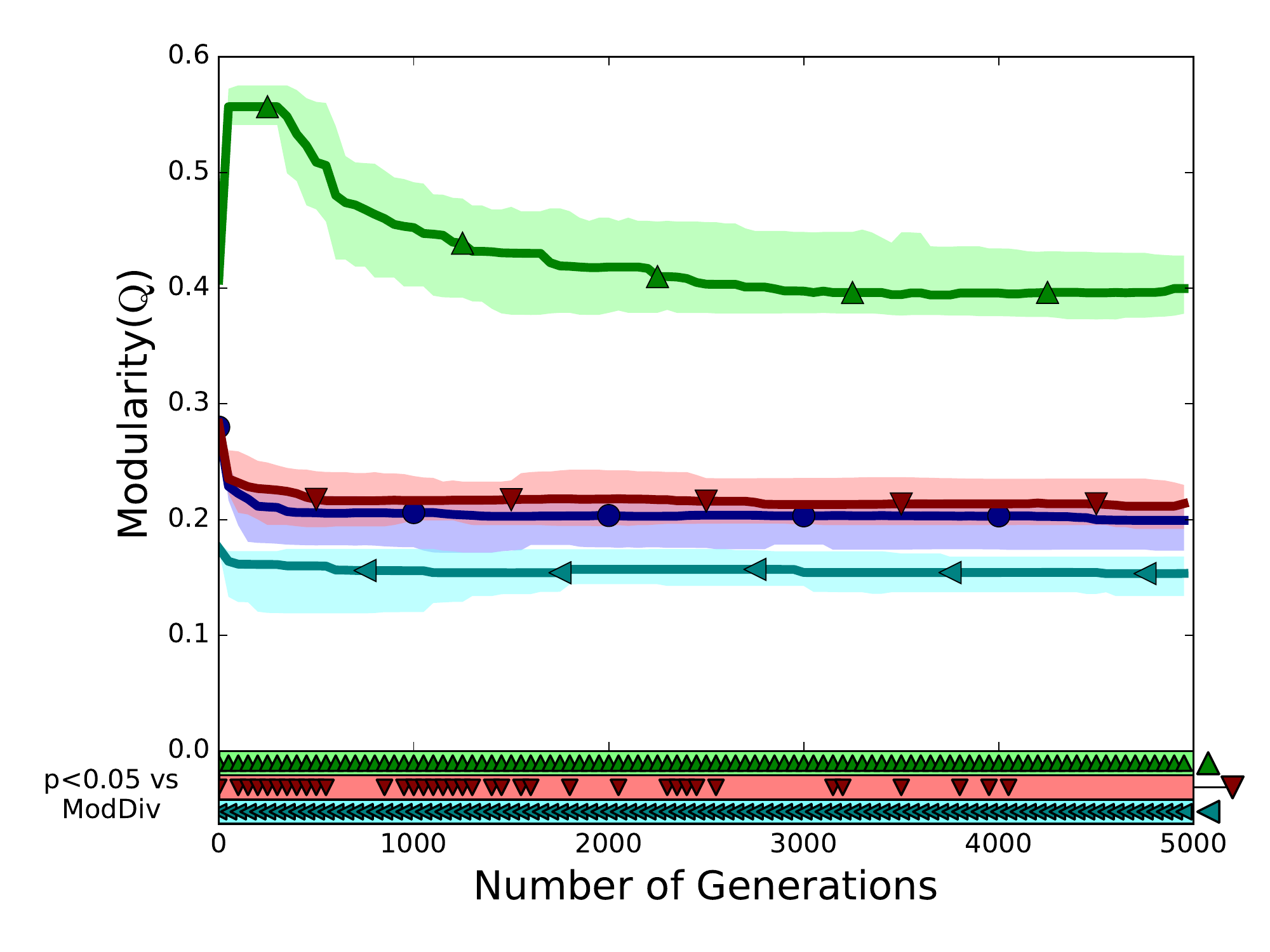}
	\caption{Median population modularity (Q-score)}
	\label{fig:nonmod_retina_modularity}
\end{subfigure}

\begin{subfigure}[b]{0.49\textwidth}
	\includegraphics[width=\textwidth]{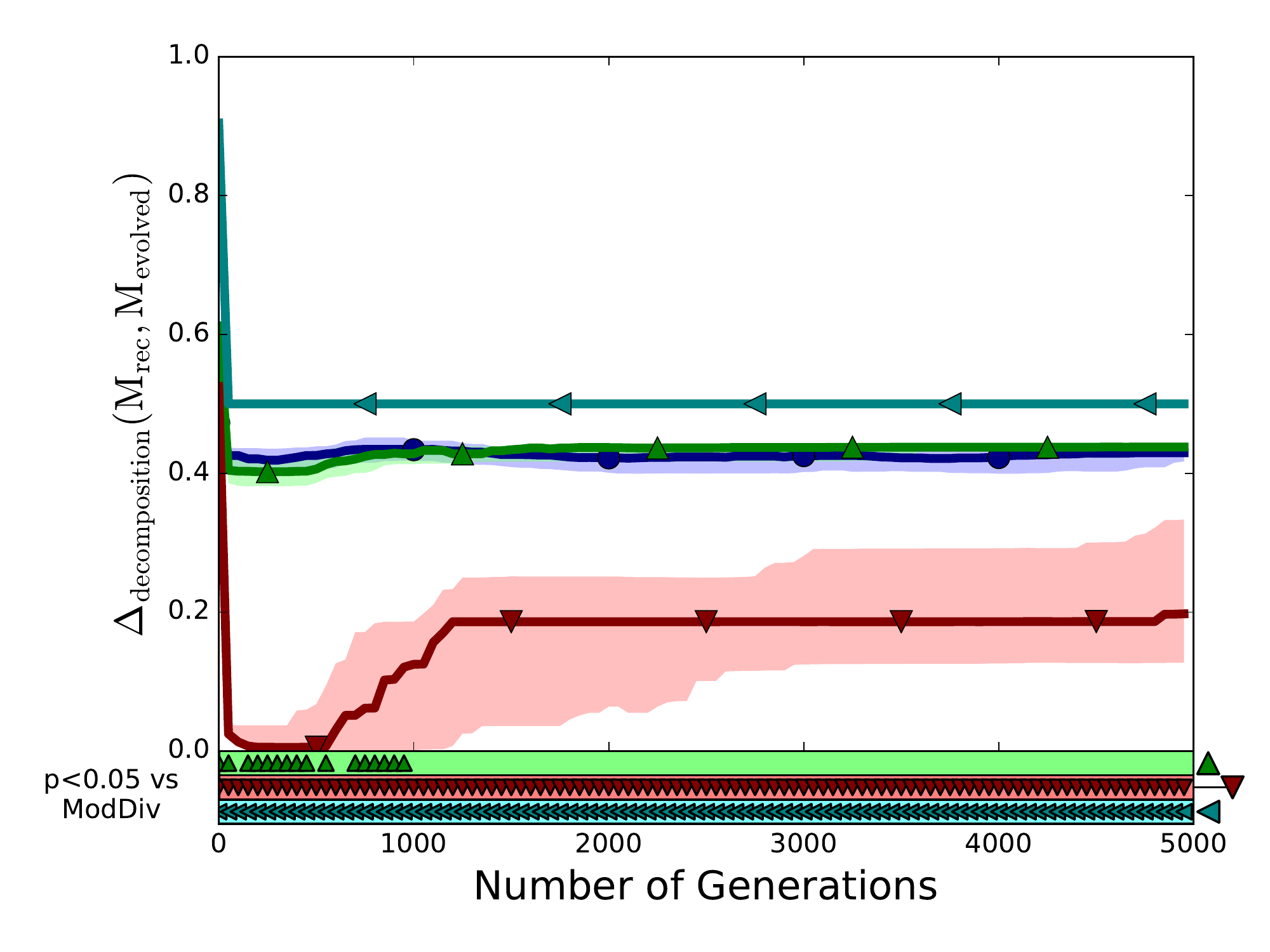}
	\caption{Median population distance from $M_{rec}$}
	\label{fig:nonmod_retina_featuremod}
\end{subfigure}
\begin{subfigure}[b]{0.49\textwidth}
	\includegraphics[width=\textwidth]{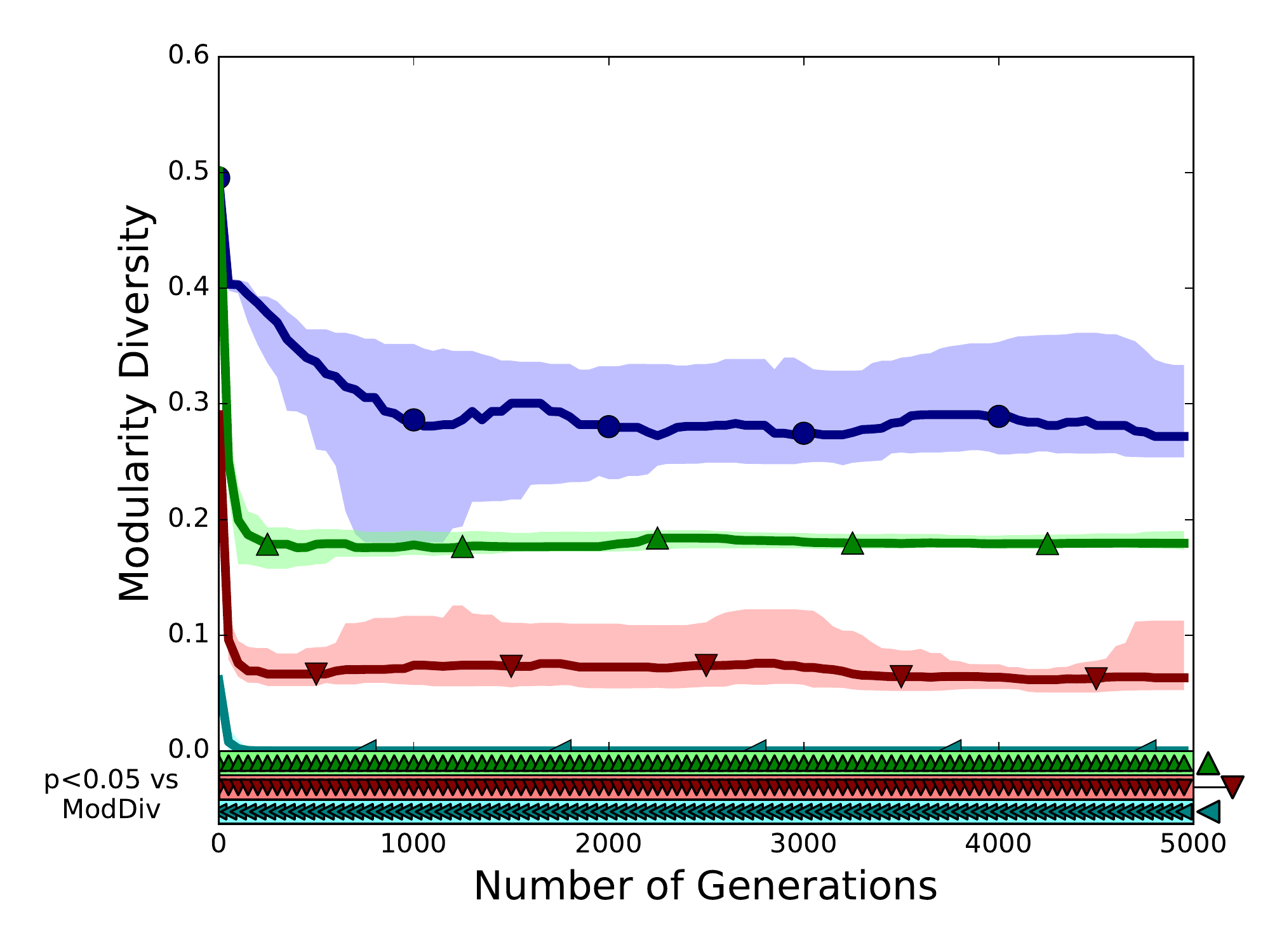}
	\caption{Median population diversity.}
	\label{fig:nonmod_retina_diversity}
\end{subfigure}
	\caption{The performance and development of different modular decompositions during evolution on the non-modular retina problem. Modularity Diversity yields the best performance, whereas the other structural objectives now result in worse performance compared to evolution guided by performance alone (PA).}
	\label{fig:nonmod_retina_result}
\end{figure}

By making all eight inputs to the retina problem a single ``pattern detector'', 
the problem becomes non-modular (Section~\ref{sec:methods_retina}). $M_{rec}$ 
for this non-modular problem was the same pattern as before 
(Figure~\ref{fig:retina_decomposition}). Since there is no modular structure in 
this problem, there is no other recommended decomposition that we expect 
to be a good guide for evolution here. As in the 
modular version, the three modularity-inducing objectives have differing 
effects with regards to the amount of modularity (Q), diversity and specific 
modularity patterns in evolved networks 
(Figure~\ref{fig:nonmod_retina_result}). As one might expect, this nonmodular 
problem no longer benefits from the guidance of the modularity-maximizing or 
User-defined Modularity objective (Figure~\ref{fig:nonmod_retina_performance}). 
However, the Modularity Diversity 
objective still improves performance significantly. Our interpretation is that 
a diverse set of high-level network structures help guide evolution, 
independently of the structure of the target problem.


\subsection{Scaling up}

One important direction for future experiments is to investigate the ability of the structural objectives to guide evolution on more complex problems, including more challenging simulated reinforcement learning problems and real-world tasks. Evolutionary algorithms have recently been demonstrated to be a viable technique for challenging reinforcement learning problems, rivaling the performance of popular backpropagation-based deep learning techniques~\citep{Petroski2017,Salimans2017}. Further, it was recently demonstrated that techniques for guiding neuroevolution by encouraging novel behaviors (originally developed for small-scale evolved networks) are also valuable when scaling up to deep reinforcement learning tasks~\citep{Conti2017a}. It is well known that the structure of deep neural networks is very important for their performance, and evolutionary algorithms are emerging as a competitive way of finding effective architectures~\citep{Real2018}. As such, it is likely that an evolutionary algorithm which searches not just for optimal performance, but which also explores many different ways of structurally organizing the network will find solutions that perform well in these deep neural networks.


Since detecting clusters of modules in a neural network is an NP-complete problem~\citep{Brandes2008}, calculating the modular decomposition of a neural network may seem like an impediment to scaling up to very large networks. However, similar to previous papers applying modularity measurements as part of neuroevolution~(e.g. \citep{Ellefsen2015, Clune2013}), we apply the spectral optimization method in our modularity calculations, which gives good results in practice at a low computational cost~\citep{Newman2006a,Fortunato2010}. Modularity calculation is only needed to be done once per network, whereas measuring the \emph{performance} of networks will usually require hundreds or thousands of passes of data through the network, as well as other computations, such as physics simulation (for robotics tasks) or training the neural network (when evolving network structures for supervised learning tasks). This performance measurement will in most cases by far be the most time consuming part of neuroevolution.


The most successful and popular application of deep learning, including deep 
reinforcement learning, is solving difficult problems directly from pixel 
inputs with deep convolutional neural networks~\citep{LeCun2015, Mnih2015a}. 
These networks already have a very specific modular structure, inspired by 
visual processing in living creatures, and are very efficient at recognizing 
objects in images~\citep{Simonyan2014}. Although evolutionary algorithms have 
been demonstrated to be a powerful technique for searching for \emph{high-level 
architectures} for convolutional neural networks~\citep{Real2018}, it is not 
likely that a neural network with a freely evolving structure (like the ones we 
study herein) would outcompete state-of-the-art convolutional networks. 
However, the networks applied in deep reinforcement learning usually have fully 
connected layers following the convolutions, which map high-level state 
representations to actions. While out of the scope of the current study, an 
intriguing opportunity is to apply structurally guided neuroevolution only to 
the latter part of the network -- using e.g. a pre-trained convolutional 
network as front-end~\citep{Poulsen2017}. Exploring different neural network 
structures here could potentially aid evolution by guiding it towards networks 
grouping together states that require similar actions.


\section{Conclusion}

We have explored the ability of objectives related to the high-level structure of neural networks to act as guiding objectives for neuroevolution. Our results are in line with previous work demonstrating that modularity-encouraging objectives can guide neuroevolution~\citep{Clune2013}, and add to that work by 1) showing that applying \emph{specific modular decompositions} as guiding objectives aids evolution on tasks with very clear, modular structure and 2) showing that guiding evolution towards a population with a \emph{diverse set of modular decompositions} increases performance both on modular and non-modular problems. This Modularity-Diversity technique is even demonstrated to produce results comparable to the powerful and popular \emph{behavioral diversity} technique.

We also demonstrated that evolution guided towards a single user-defined decomposition does not perform well for tasks that do not have a very obvious structure. This agrees with previous work demonstrating that evolving neural networks often end up with unexpected decomposition patterns not agreeing with human intuition~\citep{Huizinga2016, Schrum2016a, Ellefsen2015}. The technique of guiding evolving neural networks towards a diversity of decomposition patterns presents a way to take advantage of unexpected, creative solutions -- allowing an automatic way to discover many functional problem decompositions. The fact that the Modularity-Diversity technique showed good performance on two very different types of neural network, and genotype-phenotype mappings, strengthen our confidence that it will be a valuable asset on a large range of neuroevolution problems.

Central to these findings is our new technique for measuring the distance between modularity patterns in pairs of neural networks. A key to this technique is that it compares high-level structures of neural networks, rather than their exact patterns of connectivity. Previous work has shown such lower-level structural distance measures to be a poor guide for neuroevolution~\citep{Mouret2009a}. Presumably, low-level structure in neural networks are not very indicative of how the network decomposes and solves a problem. Thus, by applying structural comparisons on a higher level, we both maintain computational complexity within reasonable limits, and capture the \emph{most important} structural differences between networks.

This paper has presented initial evidence of the power of guiding evolution 
with high-level structural objectives, and there are many important issues to 
address for future work. First, the structural distance measure we propose is 
calculated by analyzing how \emph{input or output neurons} of networks are 
modularly separated. For some tasks, we may not expect the problem 
decomposition to be present at the input/output level, but only at intermediate 
stages of processing. In this case, we would need a way to compare high-level 
structure based on the modularity pattern of \emph{internal neurons}. This 
could be facilitated by giving each internal neuron a separate ID, and keeping 
the number of neurons at each layer of the neural network fixed (but allowing 
connections to/from them to appear and disappear). Another interesting 
direction for further research is to better understand the relationship between 
high-level structural diversity and behavioral diversity as guiding objectives. 
Our results indicate that the former may work best for problems that have a 
clearly decomposable structure which is reflected also in the structure of 
high-performing neural networks, whereas the two methods performed similarly on 
a problem with a less obvious structure. Further systematic tests of the two 
with different problem types and behavior descriptors will help uncover the 
strengths and limitations of each method.

\section{Acknowledgments}

This work is supported by The Research Council of Norway as part of the 
Engineering Predictability with Embodied Cognition (EPEC) project $\#$240862, 
the 
Collaboration on Intelligent Machines (COINMAC) project, under grant agreement 
261645 and the Centres of Excellence scheme, project $\#$262762.
The experiments were performed on the Abel Cluster, owned by the University of Oslo and the Norwegian metacenter for High Performance Computing (NOTUR), and operated by the Department for Research Computing at USIT, the University of Oslo IT-department \url{http://www.hpc.uio.no/}.
We thank Roby Velez for valuable feedback on the manuscript.

\footnotesize
\bibliographystyle{apalike}
\bibliography{bibliography} 

\end{document}